\pdfoutput=1
\documentclass[11pt]{article}

\usepackage[preprint]{acl}

\usepackage{times}
\usepackage{latexsym}

\usepackage[T1]{fontenc}
\usepackage[utf8]{inputenc}

\usepackage{microtype}

\usepackage{inconsolata}

\usepackage{graphicx}

\usepackage{tabularx}
\usepackage{xcolor}
\usepackage{booktabs}
\usepackage{colortbl}
\usepackage{svg}
\usepackage{amsmath}
\usepackage{amssymb}
\usepackage{wrapfig}
\usepackage{float}
\usepackage{algpseudocode}
\usepackage[ruled,vlined]{algorithm2e}
\usepackage{anyfontsize}
\usepackage{pgf}
\usepackage{authblk}
\usepackage{ulem}
\usepackage{listings}
\usepackage{subcaption}
\everymath=\expandafter{\the\everymath\displaystyle}
\lstdefinestyle{pythonstyle}{
    language=Python,
    basicstyle=\ttfamily\small,
    keywordstyle=\color{blue},
    stringstyle=\color{red},
    commentstyle=\color{green!60!black},
    numbers=left,
    numberstyle=\tiny,
    stepnumber=1,
    numbersep=5pt,
    backgroundcolor=\color{gray!10},
    showspaces=false,
    showstringspaces=false,
    showtabs=false,
    tabsize=2,
    captionpos=b,
    breaklines=true,
    breakatwhitespace=false,
    title=\lstname,
    frame=single,
    morekeywords={False,None,True,and,def,or,return},
    deletekeywords={type,input,print}
}

\newcommand{\squishlist}{
 \begin{list}{$\bullet$}
  { \setlength{\itemsep}{0pt}
     \setlength{\parsep}{2pt}
     \setlength{\topsep}{2pt}
     \setlength{\partopsep}{0pt}
     \setlength{\leftmargin}{0.75em}
     \setlength{\labelwidth}{0.75em}
     \setlength{\labelsep}{0.35em} } }

\newcommand{\squishend}{
  \end{list}  }

\title{TACO-RL: Task Aware Prompt Compression Optimization with Reinforcement Learning}

\author{
    \textbf{Shivam Shandilya \quad Menglin Xia \quad Supriyo Ghosh \quad Huiqiang Jiang} \\
    \textbf{Jue Zhang \quad Qianhui Wu \quad Victor Rühle} \\
    {\large Microsoft}\\
    \texttt{\{t-shandilyas, mollyxia, supriyoghosh\}@microsoft.com}
}

\begin{document}
\maketitle
\begin{abstract}
% Task aware prompt compression using RL (COLING 2025).
The increasing prevalence of large language models (LLMs) such as GPT-4 in various applications has led to a surge in the size of prompts required for optimal performance, leading to challenges in computational efficiency. Prompt compression aims to reduce the inference cost by minimizing input tokens without compromising on the task performance. However, existing prompt compression techniques either rely on sub-optimal metrics such as information entropy or model it as a task-agnostic token classification problem that fails to capture task-specific information.  

To address these issues, we propose a novel and efficient reinforcement learning (RL) based task-aware prompt compression method. To ensure low latency requirements, we leverage existing Transformer encoder-based token classification model while guiding the learning process with task-specific reward signals using lightweight REINFORCE algorithm. We evaluate the performance of our method on three diverse and challenging tasks including text summarization, question answering and code summarization. We demonstrate that our RL-guided compression method improves the task performance by 8\% - 189\% across these three scenarios over state-of-the-art compression techniques while satisfying the same compression rate and latency requirements.    

%To address these issues, we propose a novel and efficient reinforcement learning (RL)-based approach to compress prompts before they are sent to LLMs, improving efficiency without compromising task performance. Building upon existing methods that utilize encoder models for prompt compression, our approach leverages RL to optimize encoder outputs for task-specific scenarios. We specifically target tasks such as summarization, question answering (QA), and code completion, where the quality of the compressed prompt significantly impacts downstream performance. The RL policy is trained to maximize a two-fold task specific reward signal derived from the nature of the task and similarity between the outputs of the LLMs when fed with compressed and original prompts. Our method demonstrates substantial improvements in compression effectiveness and end performance across these tasks, paving the way for more scalable and resource-efficient LLM applications.
\end{abstract}

\section{Introduction}
In recent years, Large Language Models (LLMs) have experienced a surge in popularity due to their impressive performance on a wide range of natural language processing tasks \cite{brown2020languagemodelsfewshotlearners,chowdhery2022palmscalinglanguagemodeling}, ranging from question answering \cite{ushio2023empirical} to code generation \cite{chen2021evaluating} to incident management \cite{ahmed2023recommending,zhang2024automated,goel2024x}. To effectively utilize these models, various prompting techniques have been introduced, such as In-Context Learning (ICL) \cite{brown2020languagemodelsfewshotlearners}, Chain-of-Thought (CoT) \cite{wei2023chainofthoughtpromptingelicitsreasoning}, and Retrieval Augmented Generation (RAG) \cite{lewis2021retrievalaugmentedgenerationknowledgeintensivenlp}. While these techniques improve the performance and efficacy of LLMs by providing them with relevant context and guidance, these lead to increase in input prompt context length which leads to higher inference cost and latency requirements.

%However, as LLMs continue to grow in size, the computational cost and latency associated with their use also increase. This poses a significant challenge in real-world applications where low latency is crucial. 
To address this issue, several prompt compression techniques (which attempt to reduce the context length without losing essential information) have been introduced. These existing work can be broadly categorized into two major threads: (a) Task-aware compression models \cite{chuang2024learningcompresspromptnatural}, that generally finetune a task-specific decoder model that leads to high inference latency and cost; and (b) Task-agnostic compression models that either removes tokens/lexical units \cite{jiang-etal-2023-llmlingua,li-etal-2023-compressing} based on their information entropy or train a supervised binary token classification model using expert compressed examples \cite{pan2024llmlingua2datadistillationefficient}. Therefore, existing solutions either fail to capture task-specific behaviors or lead to high inference cost and latency. These challenges lead to two important research questions: 
% such as LLMLingua \cite{jiang-etal-2023-llmlingua, jiang2024longllmlinguaacceleratingenhancingllms} have been proposed. In their work, an encoder model is used to predict binary labels for input tokens, preserving tokens with a label of 1 and removing those with a label of 0.

\begin{itemize}
% \squishlist
    \item How can we design a prompt compression model that effectively leverages bidirectional context \cite{devlin2019bertpretrainingdeepbidirectional} and provides low inference latency (\textbf{Q1})?
    \item To minimize the computational cost needed for adapting this model to a new task, how can we efficiently train a model with proper guidance from task-specific reward signals (\textbf{Q2})?
% \squishend
\end{itemize}

% \setkeys{Gin}{draft}
\begin{figure*}[t]
  \centering
  \includegraphics[width=1\linewidth]{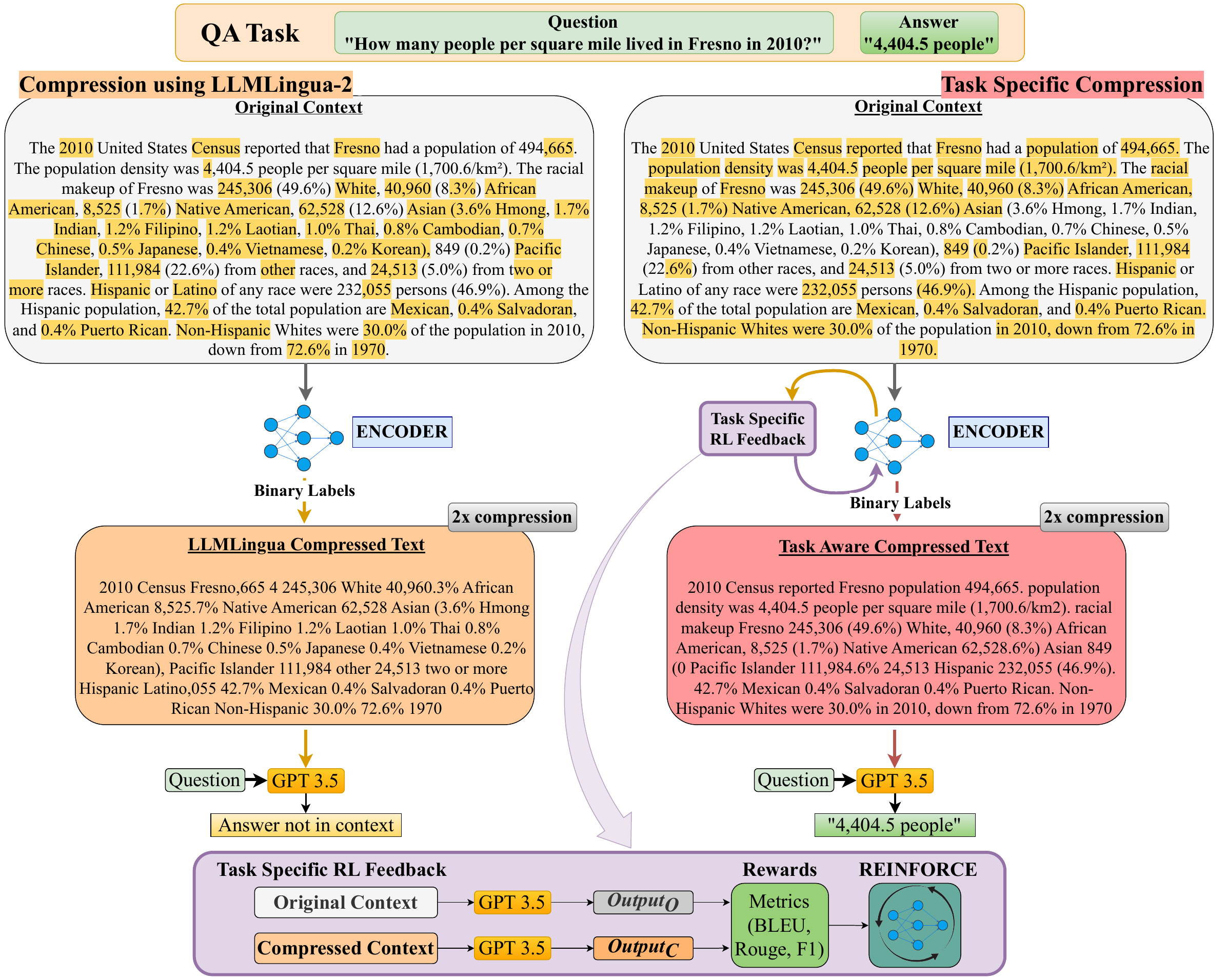}
  \caption{Encoder fine-tuning with RL using task specific reward signals on a Q/A task. The RL-guided compression model is able to understand the specificity of the question and retains the relevant context in the compressed prompt. }
  \label{fig:LLMLinguaVsOurs}
\end{figure*}

To address Q1, we build our work on the foundation laid by LLMLingua-2 \cite{pan2024llmlingua2datadistillationefficient} which trained a task-agnostic (while being inference latency-aware) encoder-based transformer model in a supervised setting for binary classification of input tokens where the target compressed prompt is generated using an expensive and efficient LLM model such as GPT-4 \cite{achiam2023gpt}. We propose a novel approach, Task-Aware Prompt Compression Optimization with Reinforcement Learning (\textbf{TACO-RL}), to guide the finetuned generic encoder model with task-specific reward signal (to address Q2) using on-policy reinforcement learning technique. As shown in Figure \ref{fig:LLMLinguaVsOurs}, during the model alignment process, we generate the task output from both the original and compressed prompt, and compute the task-specific reward signal using the divergence between these two outputs. These reward signals are then used to update the base encoder model using on-policy REINFORCE algorithm \cite{williams1992simple}. 

% Building upon the foundation laid by LLMLingua, our project aims to further improve the performance of the encoder model on task-specific scenarios. We propose a novel approach that leverages GPT-3.5 as an end reward model to fine-tune the encoder model on a given dataset. By comparing the outputs of the original and compressed prompts using task-specific metrics, we obtain rewards that are used to update the parameters of the encoder model using the REINFORCE algorithm \cite{10.1007/BF00992696}.
To illustrate the efficacy of our proposed method against state-of-the-art (SoTA) prompt compression methods, we conducted extensive experiments on three diverse and challenging tasks on open-source benchmark datasets: (a) text summarization on MeetingBank dataset; (b) question-answering tasks on Squad dataset and (c) code summarization on CodeSearchNet dataset. The empirical results across these tasks demonstrate that our RL-guided prompt compression method can improve the task performance by 8\% - 189\% over LLMLingua-2 and other benchmark approaches while ensuring same compression rate and inference cost or latency. To that end, our key contributions are as follows:
% \squishlist
\begin{itemize}
    \item We introduce a \emph{latency-aware} encoder based Transformer model for prompt compression that is aligned with task-specific objectives and leads to low inference cost. %\supriyo{Point 1 and 2 express similar view, need to reframe}
    \item We propose an efficient \emph{task-aware} prompt compression model that leverages task specific reward signals to fine-tune base language models using on-policy RL (i.e., REINFORCE algorithm).
    %-based prompt compression method that utilizes GPT-3.5 to generate end reward signals and fine-tune a encoder-based transformer model on task-specific datasets.
    \item We conduct extensive experiments on three diverse tasks to evaluate the effectiveness of our proposed method and demonstrate that our TACO-RL method provides significant improvements over SoTA methods. We will open-source the code upon publication.
\end{itemize}
% \squishend

\section{Related Work}

\paragraph{Prompt compression for LLMs.} Prompt compression shortens the input prompt to improve the inference efficiency of LLMs over long context. The form of the compressed prompts can vary, including token pruning, abstractive summarization, prompt paraphrasing, or soft prompt tuning. For example, token pruning trims less important tokens from the prompt \citep{li-etal-2023-compressing}, while abstractive summarization or prompt paraphrasing aims to condense the semantics into shorter concise text \cite{xu2024recomp}. Soft prompt tuning, on the other hand, converts the original prompt into a vector \citep{mu2024learning}. Among these methods, token pruning has proven to be particularly effective due to its flexibility and smaller computational overhead compared to other methods. Prior work on prompt pruning, such as Selective Context \citep{li-etal-2023-compressing} and LLMLingua \citep{jiang-etal-2023-llmlingua, jiang2024longllmlinguaacceleratingenhancingllms}, use heuristic metrics to compute token importance and trim less important tokens. LLMLingua-2 \citep{pan2024llmlingua2datadistillationefficient} trains a transformer-based classifier on compression data distilled from GPT-4 to decide whether to prune a token. While these task-agnostic prompt compression methods are effective and generalize to some tasks, they still struggle to model token importance in specific tasks or domains.

\paragraph{RL-based prompt compression.} RL-based methods have also been applied to prompt compression. For example, \citet{Jung_2024} leverage RL to train an MLP classifier conditioned on the task language model for token pruning. However, their compressor depends on the hidden representations of a white-box model, and the prompts they consider are usually short instructions rather than long contexts. \citet{chuang2024learningcompresspromptnatural} use RL to train a generative language model to compress prompts, which typically has a high computational overhead for compression. They consider classification tasks where there is a more straightforward signal for reward. \citet{huang2024fewermoreboostingllm} apply context pruning to in-context learning (ICL) examples, training a hierarchical pruner with RL to select more relevant ICL examples and preserve more important tokens in the selected examples for better demonstrations in mathematical reasoning. In addition to the different design choices and compression goals, our work demonstrates how we effectively combine offline and online training to obtain a task-aware prompt compression model that can better model token importance with low inference cost and latency.

\section{Background}

In this section, we provide an overview of the prompt compression and reinforcement learning (RL) framework, which are used as building blocks for our method.

% Our work takes inspiration from two key components: the prompt compression technique introduced by LLMLingua-2 \cite{pan2024llmlingua2datadistillationefficient} and the REINFORCE \cite{williams1992simple} algorithm for policy gradient optimization. Using them we propose a novel application of reinforcement learning for fine-tuning the compression model. We briefly introduce each of these components below.

\subsection{Prompt Compression}
The goal of prompt compression is to reduce the context length without losing the essential information of the original prompt. To ensure low latency and inference cost, we build our work on recently introduced 
LLMLingua-2 \cite{pan2024llmlingua2datadistillationefficient} framework that translates the problem into a binary token classification problem using an encoder-based Transformer model.

A key innovation in our approach is leveraging bidirectional context to make more informed compression decisions. Unlike unidirectional models, the encoder model captures contextual information from both left and right directions, enabling a more nuanced understanding of each token's importance. This bidirectional context is crucial in determining the relevance of a token within the broader prompt context.

Given an input prompt $X = (x_1, x_2, ..., x_N)$ with $N$ tokens, the compression process is defined as:

\begin{equation}
\mathbf{H} = \text{Encoder}(X) \in \mathbb{R}^{d \times N} \nonumber
\end{equation}
where $\mathbf{H} = [\mathbf{h}_1, \mathbf{h}_2, ..., \mathbf{h}_N]$, and $\mathbf{h}_i \in \mathbb{R}^d$ is the encoded representation of token $x_i$.

The contextual representation $\mathbf{h}_i$ captures the token's significance by considering its interactions with surrounding tokens through self-attention mechanisms. This means the probability of preserving or removing a token is not determined in isolation, but by its relationship to the entire prompt context.

For each token, a binary classification probability is computed:
\begin{equation}
    \mathbf{p}_i = \text{softmax}(\mathbf{W}\mathbf{h}_i + \mathbf{b}) \nonumber
\end{equation}
where $\mathbf{W} \in \mathbb{R}^{2 \times d}$ and $\mathbf{b} \in \mathbb{R}^2$ are learnable parameters. The compression decision $y_i$ for the \emph{i}-th token is determined by:

\begin{equation}
    y_i = \begin{cases} 
        1 & \text{if } \mathbf{p}_i[1] \geq 0.5 \\
        0 & \text{otherwise}
    \end{cases} \nonumber
\end{equation}

The compressed prompt $X_c$ is then constructed by retaining only the tokens where $y_i = 1$. The compression effectiveness is computed by Compression Rate ($\tau = |X_c| / |X|$), while the Compression Ratio ($C.R. = 1/\tau$) represents the factor by which the prompt is compressed.

\subsection{REINFORCE Algorithm}
Reinforcement learning (RL) has proven to be effective for domain alignment of language models. The seminal work of \citet{ouyang2022training} demonstrate that the learning of decoder based Transformer (e.g., GPT) models can be represented as bandit environment and off-policy RL algorithms such as Proximal Policy Optimization (PPO) \cite{schulman2017proximal} with task specific reward signals can be an effective tool for domain alignment. However, as off-policy RL is generally computationally expensive and sample inefficient, we primarily focus on guiding our compression model with on-policy RL method.  
The REINFORCE algorithm \cite{williams1992simple} is a popular on-policy policy gradient method that can be used to optimize a parameterized policy. Let $\pi_\theta(a|s)$ denote a parameterized policy with parameters $\theta$ that given a state $s$, can generate the probability of executing action $a$. Our goal is to optimize the parameters $\theta$ to maximize the expected return: $J(\theta) = \mathbb{E}_{\tau \sim \pi_\theta}[R(\tau)] $,
% \begin{equation}
%     J(\theta) = \mathbb{E}_{\tau \sim \pi_\theta}[R(\tau)] \nonumber
% \end{equation}
where $R(\tau)$ denotes the cumulative rewards obtained for trajectory $\tau$. The gradients used to update the policy parameters are computed using Eq.~\ref{rl_gradient}.
\begin{equation}
    \nabla_\theta J(\theta) = \mathbb{E}_{\tau \sim \pi_\theta}[\nabla_\theta \log \pi_\theta(a|s) R(\tau)] \label{rl_gradient}
\end{equation}

% \subsection{Applying REINFORCE to Prompt Compression}

% In our approach, we cast the prompt compression task as a reinforcement learning problem. The encoder model serves as the policy $\pi_\theta$, where $\theta$ represents the encoder's parameters. The state $s$ is the input prompt, and the actions $a$ are the binary decisions to keep or discard each token.

% We use GPT-3.5 as an end reward model to evaluate the quality of the compressed prompts. The reward $R(\tau)$ is computed by comparing the outputs of GPT-3.5 for both the original and compressed prompts using task-specific metrics.

% The REINFORCE algorithm is then applied to update the encoder's parameters:

% \begin{equation}
%     \theta \leftarrow \theta + \alpha \nabla_\theta \log \pi_\theta(a|s) R(\tau)
% \end{equation}
% where $\alpha$ is the learning rate. This process allows us to fine-tune the encoder model for task-specific prompt compression, potentially improving its performance beyond the general compression technique of LLMLingua-2.

% By combining these components, our work aims to create a more adaptive and task-specific prompt compression method, leveraging the power of reinforcement learning to optimize the compression process for particular applications in natural language processing.

\begin{table*}[!ht]
  \centering
  \resizebox{1\textwidth}{!}{
    \begin{tabular}{l>{\raggedleft\arraybackslash}c>{\raggedleft\arraybackslash}c>{\raggedleft\arraybackslash}c>{\raggedleft\arraybackslash}c>{\raggedleft\arraybackslash}c}
        \toprule
        \textbf{Models} & \textbf{Bleu} & \textbf{Rouge1} & \textbf{Rouge2} & \textbf{RougeL} & \textbf{BertScore F1} \\
        \midrule
        \multicolumn{6}{c}{\textbf{0.50 (2x compression)}} \\
        \hline
        LLMLingua-2 - MeetingBank & 18.68 & 54.20 & 29.45 & 40.14 & 90.69 \\
        LLMLingua-2 - Wikitext & 16.71 (-1.97) & 52.58 (-1.62) & 27.73 (-1.72) & 39.05 (-1.09) & 90.47 (-0.21) \\
        LLMLingua & 5.90 (-12.78) & 38.22 (-15.98) & 14.02 (-15.42) & 26.49 (-13.65) & 87.65 (-3.04) \\
        Selective Context & 12.94 (-5.74) & 46.30 (-7.90) & 24.41 (-5.03) & 34.56 (-5.58) & 89.66 (-1.03) \\
        \rowcolor{lightgray} \textbf{TACO-RL (Ours)} & \textbf{21.35 (+2.67)} & \textbf{55.34 (+1.14)} & \textbf{31.88 (+2.43)} & \textbf{42.17 (+2.03)} & \textbf{90.95 (+0.26)} \\
        \hline
        \multicolumn{6}{c}{\textbf{0.33 (3x compression)}} \\
        \hline
        LLMLingua-2 - MeetingBank & 15.11 & 51.67 & 25.60 & 37.18 & 90.17 \\
        LLMLingua-2 - Wikitext & 12.93 (-2.18) & 49.38 (-2.29) & 23.32 (-2.28) & 35.34 (-1.83) & 89.79 (-0.38) \\
        LLMLingua & 3.98 (-11.13) & 32.62 (-19.05) & 10.58 (-15.02) & 22.58 (-14.60) & 86.52 (-3.65) \\
        Selective Context & 8.80 (-6.31) & 40.22 (-11.45) & 19.28 (-6.32) & 29.44 (-7.74) & 88.67 (-1.50) \\
        \rowcolor{lightgray} \textbf{TACO-RL (Ours)} & \textbf{19.36 (+4.26)} & \textbf{53.67 (+1.99)} & \textbf{29.54 (+3.94)} & \textbf{40.01 (+2.83)} & \textbf{90.54 (+0.37)} \\
        \hline
        \multicolumn{6}{c}{\textbf{0.25 (4x compression)}} \\
        \hline
        LLMLingua-2 - MeetingBank & 12.80 & 49.40 & 22.77 & 34.77 & 89.78 \\
        LLMLingua-2 - Wikitext & 10.98 (-1.82) & 47.34 (-2.07) & 20.68 (-2.09) & 33.06 (-1.71) & 89.34 (-0.44) \\
        LLMLingua & 3.51 (-9.29) & 30.98 (-18.42) & 9.64 (-13.13) & 21.33 (-13.45) & 86.20 (-3.58) \\
        Selective Context & 6.37 (-6.43) & 36.22 (-13.18) & 15.97 (-6.80) & 26.07 (-8.71) & 88.00 (-1.78) \\
        \rowcolor{lightgray} \textbf{TACO-RL (Ours)} & \textbf{17.61 (+4.81)} & \textbf{52.33 (+2.92)} & \textbf{27.84 (+5.07)} & \textbf{38.57 (+3.79)} & \textbf{90.26 (+0.48)} \\
        \hline
        \multicolumn{6}{c}{\textbf{0.20 (5x compression)}} \\
        \hline
        LLMLingua-2 - MeetingBank & 11.13 & 47.50 & 21.01 & 33.25 & 89.44 \\
        LLMLingua-2 - Wikitext & 9.51 (-1.61) & 45.33 (-2.17) & 18.71 (-2.30) & 31.27 (-1.98) & 88.90 (-0.54) \\
        LLMLingua & 3.42 (-7.71) & 30.53 (-16.98) & 9.62 (-11.38) & 21.15 (-12.09) & 86.14 (-3.29) \\
        Selective Context & 4.82 (-6.31) & 33.15 (-14.35) & 13.55 (-7.46) & 23.74 (-9.51) & 87.49 (-1.94) \\
        \rowcolor{lightgray} \textbf{TACO-RL (Ours)} & \textbf{15.85 (+4.73)} & \textbf{50.56 (+3.06)} & \textbf{26.04 (+5.03)} & \textbf{36.81 (+3.56)} & \textbf{89.96 (+0.52)} \\
        \hline
        \multicolumn{6}{c}{\textbf{0.166 (6x compression)}} \\
        \hline
        LLMLingua-2 - MeetingBank & 9.80 & 45.82 & 19.19 & 31.64 & 89.12 \\
        LLMLingua-2 - Wikitext & 8.55 (-1.25) & 43.44 (-2.38) & 17.06 (-2.13) & 29.67 (-1.96) & 88.53 (-0.59) \\
        LLMLingua & 3.19 (-6.62) & 29.85 (-15.97) & 9.47 (-9.72) & 20.75 (-10.89) & 86.07 (-3.05) \\
        Selective Context & 4.06 (-5.74) & 31.34 (-14.48) & 12.21 (-6.97) & 22.31 (-9.33) & 87.12 (-2.00) \\
        \rowcolor{lightgray} \textbf{TACO-RL (Ours)} & \textbf{14.25 (+4.45)} & \textbf{48.60 (+2.78)} & \textbf{24.51 (+5.33)} & \textbf{35.08 (+3.44)} & \textbf{89.68 (+0.56)} \\
        \hline
        \textbf{Results with Original Prompts} & \textbf{21.50} & \textbf{55.19} & \textbf{33.03} & \textbf{42.90} & \textbf{91.12} \\
        \bottomrule
        \end{tabular}
    }
    \caption{\label{meetingbank_results} Performance metrics for different models across various compression rates ($\tau$) on the \textbf{MeetingBank Dataset}. Values in parentheses indicate deltas from the original LLMLingua-2 baseline.
    }
\end{table*}

\section{Methodology}
%In our approach, we cast the prompt compression task as a reinforcement learning problem. The encoder model serves as the policy $\pi_\theta$, where $\theta$ represents the encoder's parameters. The state $s$ is the input prompt, and the actions $a$ are the binary decisions to keep or discard each token.

% We use GPT-3.5 as an end reward model to evaluate the quality of the compressed prompts. The reward $R(\tau)$ is computed by comparing the outputs of GPT-3.5 for both the original and compressed prompts using task-specific metrics.

% The REINFORCE algorithm is then applied to update the encoder's parameters:

% \begin{equation}
%     \theta \leftarrow \theta + \alpha \nabla_\theta \log \pi_\theta(a|s) R(\tau)
% \end{equation}
% where $\alpha$ is the learning rate. This process allows us to fine-tune the encoder model for task-specific prompt compression, potentially improving its performance beyond the general compression technique of LLMLingua-2.

% By combining these components, our work aims to create a more adaptive and task-specific prompt compression method, leveraging the power of reinforcement learning to optimize the compression process for particular applications in natural language processing.
In this work, we combine the power of latency aware encoder-based Transformer model for token classification and on-policy RL algorithm for developing an efficient task-aware prompt compression model. We now present our proposed method Task-Aware Prompt Compression Optimization with Reinforcement Learning (TACO-RL).

\subsection{TACO-RL}
Our proposed TACO-RL framework has 3 key components: (1) A base encoder policy for action sampling and token classification, (2) Task specific reward signal computation; and (3) Policy optimization using on-policy RL method.  
%We propose a latency and task-aware compression framework that leverages reinforcement learning (RL) to optimize an encoder model for prompt compression in task-specific scenarios. Our approach aligns the compression process with the given task by using task-specific rewards to guide the learning process.

\paragraph{Encoder Model and Action Sampling.}
Given an input prompt sequence $\mathbf{x} = (x_1, \ldots, x_n)$, our encoder policy predicts probabilities $\mathbf{p} = (p_1, \ldots, p_n)$, where $p_i = P(a_i = 1 \mid x_i)$ represents the probability of preserving token $x_i$. We sample binary actions $\mathbf{a} = (a_1, \ldots, a_n)$ from these probabilities:

\begin{equation}
    a_i \sim \text{Bernoulli}(p_i) \quad \forall i \in \{ 1, \ldots, n \}
\end{equation}

The compressed prompt $\mathbf{x}_c$ is then constructed by retaining only the tokens where $a_i = 1$.

\paragraph{Reward Calculation.}
We leverage an effective but relatively cheaper LLM model, GPT-3.5-Turbo model to generate outputs $y_{\text{orig}}$ and $y_{\text{comp}}$ from the original and compressed prompts, respectively. The reward $r$ is then computed based on task-specific metrics $\mathcal{M}$ (see Section \ref{tacro_reward}):

\begin{equation}
\label{reward_formulation}
r = 
\begin{cases}
    \mathcal{M}(y_{\text{comp}}, y_{\text{orig}}), & \text{if } -L \leq \delta < L \\
    r_0, & \text{otherwise}
\end{cases}
\end{equation}

Here, $\mathcal{M}(y_{\text{comp}}, y_{\text{orig}})$ denotes the divergence metric between the output of original and compressed prompts. $r_0$ is a negative constant reward for out-of-range compression. $\delta (= |\mathbf{x}_c| - c \cdot |\mathbf{x}|)$ denotes the divergence from expected compression. We use two important parameters to control the learning process, a compression flexibility controller $c$ and a tolerance threshold $L$.

\textit{Compression Flexibility Controller.}
The compression flexibility controller $c$ is a tunable hyperparameter that represents a baseline proportion for the number of tokens to be retained in the compressed prompt relative to the original prompt. A smaller value of $c$ enforces stricter compression (fewer tokens to be retained), while a larger value allows more tokens to be preserved. This provides flexibility in controlling the trade-off between output quality and inference cost. Also, this parameter is only required to guide the encoder during training and is different from the compression ratio used during inference. We show that the a model trained with a single vale for $c$, generalizes to different compression ratios during training. %By adjusting $c$, we can fine-tune the trade-off between aggressive compression and preserving critical information, adapting the model's behavior to specific task requirements or computational constraints.

\textit{Tolerance Threshold.}
As our policy is unconstrained, the action sampling may not always satisfy the compression ratio requirements during training process. Therefore, we define tolerance threshold parameter, $L$ to control the divergence from expected number of compressed tokens. Let $\delta$ denote the divergence between the actual and expected number of compressed tokens. To ensure a smooth learning process, we allow the divergence value $\delta$ to fall within the range $[-L, L]$ and if this criteria is met, then the task-specific positive reward signal is propagated. However, to ensure that the compressed prompt is neither excessively short nor unnecessarily long, a constant negative reward $r_0$ is applied to penalize extreme deviation. This mechanism stabilizes the compression process and guides the model towards generating prompts of desired lengths. It should be noted that during inference, we always satisfy the exact compression rate by sampling $\tau\cdot|x|$ tokens with highest probability. 

% \begin{equation}
%     
% \end{equation}

\paragraph{Policy Optimization.}
Finally, with the task-specific reward signals $r$, we update the encoder policy parameters using the REINFORCE \cite{williams1992simple} algorithm. The loss function is defined as:
\begin{equation}
    \mathcal{L} = -r \sum_{i=1}^n \log p(a_i \mid x_i) - \lambda H(\mathbf{p})
\end{equation}
where $H(\mathbf{p})$ represents Shannon's entropy regularization \cite{6773024} and $\lambda$ balances the tradeoff between reward and exploration. The gradient of the loss function with respect to the model parameters $\theta$ is calculated using Eq.~\ref{taco_rl}:
\begin{equation}
\begin{split}
    \nabla_{\theta} \mathcal{L} = -\mathbb{E}_{a \sim p(\cdot \mid x; \theta)} \left[ r \nabla_{\theta} \log p(a_i \mid x_i; \theta) \right]
    \\ - \lambda \nabla_{\theta} H(\mathbf{p})
\end{split}
\label{taco_rl}
\end{equation}

By iteratively optimizing this objective, our approach refines the encoder's ability to generate compressed prompts that retain essential task-specific information while minimizing prompt length.% The interplay between the compression flexibility controller $c$ and the tolerance threshold $L$ allows for fine-grained control over the compression process, enabling the model to adapt to various task requirements and constraints.

\paragraph{Overall Approach.} Algorithm~\ref{algo1} describes the key steps of TACO-RL. We begin with a task-agnostic encoder-based Transformer model \cite{pan2024llmlingua2datadistillationefficient} for token classification. In each epoch, we generate the compressed prompt for every training data from the current encoder policy $\pi_\theta$ and compute the output of both original and compressed prompt using GPT-3.5-turbo model. Based on these outputs, we compute the task-specific output divergence metrics and use that as a positive reinforcement if the compression ratio requirements are met, otherwise we provide a small negative reward to penalize the constraint violation. Finally, using this reward signal we update the current policy parameter $\theta$ using Eq. \ref{taco_rl}. We execute this policy optimization process for $E$ epochs to generate the final optimized task-aware compression model $\pi^*_\theta$.

\begin{algorithm}
\DontPrintSemicolon
\caption{TACO-RL}
\label{algo1}
\textbf{Input:} Training set $\mathcal{D}$, initial encoder $\pi_\theta$, compression controller $c$, tolerance $L$, number of epochs $E$, Metric $\mathcal{M}$\;
\textbf{Output:} {Optimized encoder policy $\pi_\theta^*$}\;
\For{epoch $= 1$ \KwTo $E$}{
    \For{$P \in \mathcal{D}$}{
        $\mathbf{H} \leftarrow \pi_\theta(P)$\;
        \For{$w_i \in P$}{
            $p_i \leftarrow \text{softmax}(\mathbf{W}\mathbf{h}_i + \mathbf{b})$\;
            $a_i \sim \text{Bernoulli}(p_i)$\;
        }
        $P_c \leftarrow \{w_i | a_i = 1\}$\;
        $y_{\text{orig}}, y_{\text{comp}} \leftarrow \text{GPT}(P), \text{GPT}(P_c)$\;
        $\delta \leftarrow |P_c| - c|P|$\;
        % \resizebox{0.75\linewidth}{!}{
        $r \leftarrow \begin{cases}
            \text{$\mathcal{M}$}(y_{\text{comp}}, y_{\text{orig}}), & \text{if } |\delta| \leq L \\
            r_0, & \text{otherwise}
        \end{cases}$\;
        % }
        $\mathcal{L} \leftarrow -r \sum_{i} \log p(a_i | w_i) - \lambda H(\mathbf{p})$\;
        Compute $\nabla_{\theta} \mathcal{L}$\ using Eq.~\ref{taco_rl}\;
        $\theta \leftarrow \theta$ + $\nabla_{\theta} \mathcal{L}$\ ;
    }
}
\KwRet $\pi_\theta^*$    \tcp*{updated policy}
\end{algorithm}

\subsection{Task-specific Rewards} \label{tacro_reward}

This section describes the task-specific reward formulations for summarization and question answering tasks.

\subsubsection{Reward Formulation}

For both task types, we generate outputs using the original prompt $\mathbf{x}$ and the compressed prompt $\mathbf{x}_c$ through a LLM (e.g., GPT-3.5-turbo):

\[
y_{\text{orig}} = \text{GPT}(\mathbf{x}, [q]), y_{\text{comp}} = \text{GPT}(\mathbf{x}_c, [q])
\]
where $q$ is the question for QA tasks.

% \subsection{Generalized Reward Structure}

The divergence metric $\mathcal{M}(y_{\text{comp}}, y_{\text{orig}})$ in Equation \ref{reward_formulation} defines a generalized reward structure that can be applied to any task. However, the optimal reward signal may vary depending on the task. For example, for summarization, we would like to maximize the similarity between the two versions of summaries while minimizing hallucination in the summary from the compressed prompt. For question answering, the focus is on optimizing the accuracy and completeness of the generated answers.

% \begin{equation}
%     r_{\text{Task}} = \text{Similarity}(o_{\text{comp}}, o_{\text{orig}})
% \end{equation}
% where $o_{\text{comp}}$ and $o_{\text{orig}}$ are the outputs from compressed and original prompts, respectively, and $\text{Similarity}(\cdot, \cdot)$ is a task-appropriate similarity metric.

% This generalized structure guides the selection of specific metrics for each task. The key is to choose a similarity function that captures the essential aspects of the task at hand. For summarization, this involves measuring content overlap and preservation of key information. For question answering, the focus is on the accuracy and completeness of the generated answers.

% \subsection{Reward Choices}

Based on the generalized structure, we implement the following task-specific rewards:

\subsubsection{Summarization Tasks}

For summarization, we utilize the BLEU \cite{papineni-etal-2002-bleu} score:

\begin{equation}
    \mathcal{M}_{\text{Sum}} = \text{BLEU}(y_{\text{comp}}, y_{\text{orig}})
\end{equation}

As a precision-based verbatim similarity metric, BLEU effectively captures n-gram overlap between the summaries of the original and compressed contexts to promote content similarity, and minimizes introduction of inaccuracies and hallucinations. BLEU's inclusion of a penalty for extra tokens helps prevent gaming of the reward system \cite{skalse2022definingcharacterizingrewardhacking}, unlike recall-based metrics such as Rouge which could be exploited by simply generating longer outputs. This design choice results in more stable training through better-calibrated rewards.

\subsubsection{Question Answering Task}

For the QA task, we define the following precision and recall functions to measure the accuracy of the textual answer compared to the original, and we use the F1 score as the final divergence metric:

\begin{align}
    \text{Precision} &= \frac{|\{y_i | y_i \in y_{\text{orig}} \cap y_{\text{comp}}\}|}{|\{y_i | y_i \in y_{\text{comp}}\}|}\\
    \text{Recall} &= \frac{|\{y_i | y_i \in y_{\text{orig}} \cap y_{\text{comp}}\}|}{|\{y_i | y_i \in y_{\text{orig}}\}|}\\
    \mathcal{M}_{\text{QA}} &= \text{F1}(y_{\text{comp}}, y_{\text{orig}})
\end{align}

The F1 score balances precision and recall, helping to ensure that the compressed prompt retains essential context for accurate answer generation.

% These reward formulations, derived from the generalized structure, encourage the model to compress prompts while preserving task-critical information, as indicated by higher similarity scores between outputs from original and compressed prompts.

\begin{table*}[ht]
  \centering
  \large
  \renewcommand{\arraystretch}{1.2}
  \resizebox{\textwidth}{!}{
    \begin{tabular}{l>{\raggedleft\arraybackslash}c>{\raggedleft\arraybackslash}c>{\raggedleft\arraybackslash}c}
        \toprule
        \textbf{Models} & \textbf{Bleu} & \textbf{Rouge1} & \textbf{RougeL} \\
        \midrule
        LLMLingua-2 - MeetingBank & 18.68 & 54.20 & 40.14 \\
        TACO-RL - with Rouge1 & 19.89 & 54.40 & 41.22 \\
        TACO-RL - with RougeL & 19.72 & 54.11 & 41.20  \\
        \rowcolor{lightgray} \textbf{TACO-RL - with BLEU} & \textbf{21.35} & \textbf{55.34} & \textbf{42.17} \\
        \hline
        \textbf{Results with Original Prompts} & \textbf{21.50} & \textbf{55.19} & \textbf{42.90} \\
        \bottomrule
        \end{tabular}
    % }
    % \resizebox{\columnwidth}{!}{
    \begin{tabular}{l>{\raggedleft\arraybackslash}c>{\raggedleft\arraybackslash}c>{\raggedleft\arraybackslash}c}
    \toprule
    \textbf{Models} & \textbf{QA F1 Score} & \textbf{EM Score}\\
    \midrule
    LLMLingua-2 - Squad & 62.70  & 38.02  \\
    TACO-RL - with Token Wise Score & 56.67 & 35.14  \\
    TACO-RL - with F1 + Token Wise Score & 68.03 & 44.64  \\
    \rowcolor{lightgray} \textbf{TACO-RL - with F1} & \textbf{69.62} & \textbf{46.32} \\
    \hline
    \textbf{Results with Original Prompts} & \textbf{71.40} & \textbf{47.49} \\
    \bottomrule
    \end{tabular}
    }
\caption{\label{squad_ablation_scores} Ablation study on the effect of different rewards on model performance at \textbf{2x} compression rate on the \textbf{MeetingBank dataset} (left) and on the \textbf{SQuAD dataset} (right).}
\end{table*}

% \setkeys{Gin}{draft}
% \begin{figure*}[t]
%   \centering
%   \includegraphics[width=\textwidth]{Figures/squad_scores.pdf}
%   \caption{Comparison of QA F1 Scores and EM Counts across different compression rates for various models on the \textbf{Squad Dataset}. The bars represent QA F1 Scores, and the lines represent EM Counts. The two lines on top represent the scores with Original Context.}
%   \label{fig:Squad_Scores}
% \end{figure*}

% \setkeys{Gin}{draft}
% \begin{figure}[t]
%     \centering
%     \scalebox{0.5}{\input{Figures/squad_scores.pgf}}
%   \caption{Comparison of QA F1 Scores and EM Counts across different compression rates for various models on the \textbf{Squad Dataset}. The bars represent QA F1 Scores, and the dashed lines represent EM Scores. The numbers on top of the bars represent the EM Counts. The two lines on top represent the scores with Original Context.}
%   \label{fig:Squad_Scores}
% \end{figure}

\section{Experiments}
%%%% Experiments %%%%%%%%%%%%%%
Our experimentation spans three different domains, each presenting unique compression challenges. 

\subsection{Datasets and Task Diversity}

\textbf{MeetingBank} \cite{hu2023meetingbankbenchmarkdatasetmeeting} is a conversational summarization dataset with $\sim$44k train examples and 862 test samples. The dataset challenges our approach by requiring compression of complex, non-linear dialogue transcripts where contextual relevance depends on nuanced speaker interactions.

\textbf{SQuAD 2.0} \cite{rajpurkar2016squad100000questionsmachine} presents a more sophisticated question-answering challenge. Unlike straightforward summarization, this dataset requires selective information extraction where compression must preserve only the most relevant context for a specific question. Using a representative subset of $\sim$34k training and 6k test examples, the dataset tests the model's ability to perform \textit{context-aware} compression. This task is particularly challenging because different questions may require preserving entirely different segments of the same context.

\textbf{CodeSearchNet} \cite{husain2020codesearchnetchallengeevaluatingstate} presents a unique non-natural language compression challenge in code documentation. From the original large-scale dataset, we curated a subset of $\sim$25k training and 1300 test examples focusing on Python code summarization. Unlike natural language, code compression demands preserving complex syntactic structures, algorithmic logic, and domain-specific semantic nuances. The challenge lies in distilling code context into meaningful summaries by identifying key functional components, understanding the overall structure and logic flow, and retaining critical variable names and comments that convey essential information for comprehensibility.

By selecting datasets spanning summarization, question-answering, and technical documentation, we comprehensively assess our prompt compression technique. Each dataset represents a distinct real-world information compression scenario: conversational summarization, targeted information extraction, and technical context distillation.

% For more experimental details, see Appendix \ref{appendix:A}.

\subsection{Base Model Training}
Our training process involves two stages: (1) training a base model using a technique similar to LLMLingua-2 \cite{pan2024llmlingua2datadistillationefficient}, and (2) fine-tuning this base model using our novel reinforcement learning approach with task-specific rewards.

For each target dataset, we trained a base model on a distinct dataset with a similar distribution to ensure domain relevance while avoiding direct overlap. Specifically, for MeetingBank and SQuAD 2.0, we utilized the Wikitext dataset \cite{merity2016pointersentinelmixturemodels} for base model training. In the case of CodeSearchNet \cite{husain2020codesearchnetchallengeevaluatingstate}, we employed the Py150 dataset \cite{10.1145/3022671.2984041}.
To train our base model, we followed the LLMLingua-2 approach and created annotated datasets using GPT-4. Across all experiments, the base models were consistently trained for 10 epochs and the values of other hyperparameters were set the same. 
%Refer to Appendix \ref{appendix:A} for details.

\subsection{Experimental Setup}

\begin{figure*}[t]
    \centering
    \scalebox{0.9}{\input{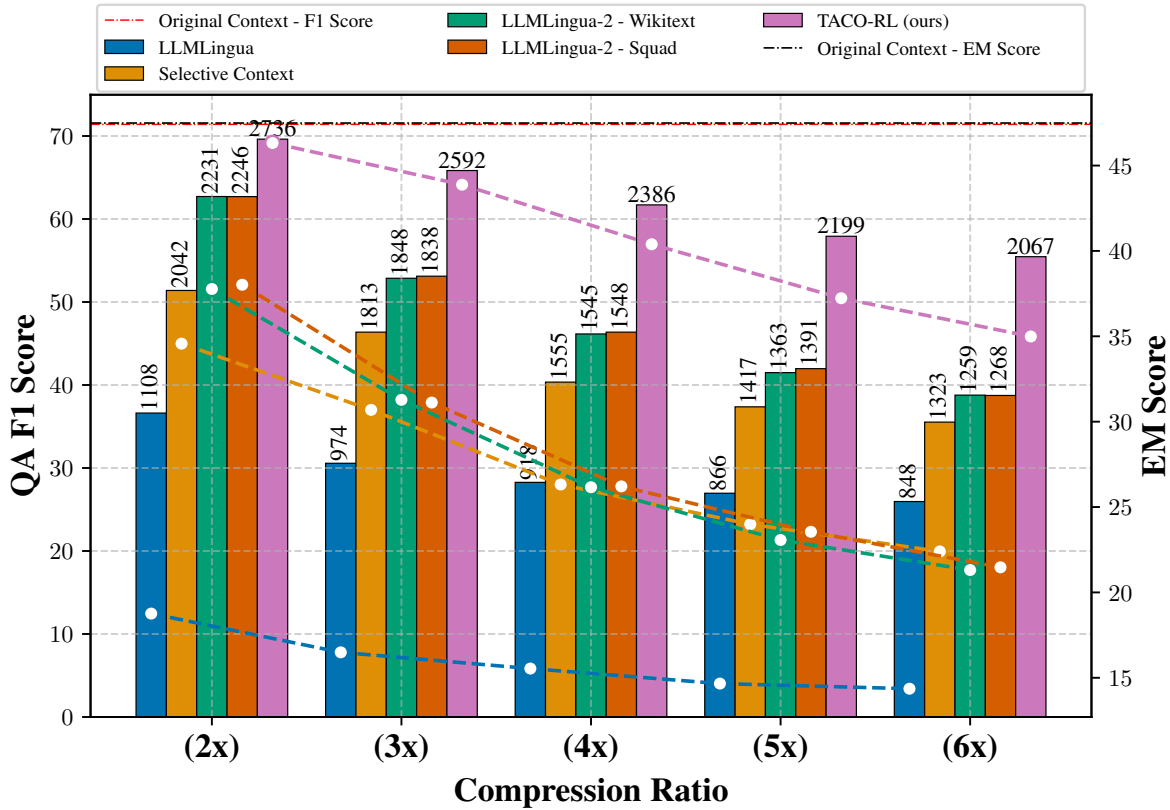}}
  \caption{Comparison of QA F1 Scores and EM Counts across different compression rates for various models on the \textbf{Squad Dataset}. The bars represent QA F1 Scores, and the dashed lines represent EM Scores. The numbers on top of the bars represent the EM Counts. The two lines on top represent the scores with Original Context.}
  \label{fig:squad_scores}
\end{figure*}

In our experiments, we employed the same architecture used in LLMLingua-2, {\fontfamily{qcr}\selectfont
xlm-roberta-large} \cite{conneau-etal-2020-unsupervised} with 561M parameters, as the backbone for our prompt compression technique. We replace the lm\_head with a classification head on top. The experiments were performed on a compute instance equipped with 8 NVIDIA V100 GPUs (32 GB variants). For downstream evaluation of the compressed prompts, we utilized GPT-3.5-Turbo-1103 as our target language model. We fixed the temperature parameter at zero to guarantee reproducible outcomes and uniform results in our experiments. During the fine-tuning phase, we employed a learning rate of 1e-6 in conjunction with a Cosine Annealing \cite{loshchilov2017sgdrstochasticgradientdescent} scheduler to stabilize the training process. Further details regarding the experimental setup can be found in Appendix \ref{appendix:A}.

%%%%%%%%%%%%%%%%%%%%%%%%%%%%%%%

% \setkeys{Gin}{draft}
% \begin{figure*}[t]
%   \centering
%   \includegraphics[width=\textwidth]{Figures/csn_scores.pdf}
%   \caption{Comparison of BLEU score across different compression rates for various models on the \textbf{CodeSearchNet} dataset.}
%   \label{fig:CSN Scores}
% \end{figure*}

% \begin{figure*}[t]
%     \centering
%     \scalebox{0.45}{\input{Figures/squad_scores.pgf}}
%     \scalebox{0.45}{\input{Figures/csn_scores.pgf}}
%   \caption{(Left) Comparison of QA F1 Scores and EM Counts across different compression rates for various models on the \textbf{Squad Dataset}. The bars represent QA F1 Scores, and the dashed lines represent EM Scores. The numbers on top of the bars represent the EM Counts. The two lines on top represent the scores with Original Context. (Right) Comparison of BLEU score across different compression rates for various models on the \textbf{CodeSearchNet} dataset.}
%   \label{fig:squad_csn_combined_scores}
% \end{figure*}

%%%%%%%%%%%%%%%%%%%%%%%%%%%%%%%%%%%%%%%%%%

\subsection{Baselines}
We use a LLMlingua-2 \cite{pan2024llmlingua2datadistillationefficient} base model trained on the same datasets as the primary baseline for all our experiments. 
%This base model is trained on the annotated version of the dataset created using GPT-4. 
We also compare our method with two other state-of-the-art compression techniques: Selective Context \cite{li-etal-2023-compressing} and LLMLingua \cite{jiang-etal-2023-llmlingua}.

\subsection{Empirical Results}
We conduct experiments on the three datasets over five different compression ratios. 
%For the two summarization tasks, we also train the encoder using different rewards. 
We observe that models trained at one choice of hyper parameters (see Table \ref{tab:hyper-parameters} in the appendix) generalize well over all compression ratios. We also conduct statistical significance tests on MeetingBank dataset results (see Table \ref{tab:sig_test_table}) to support our performance enhancement claims.

\begin{figure*}[t]
    \centering
    \scalebox{0.9}{\input{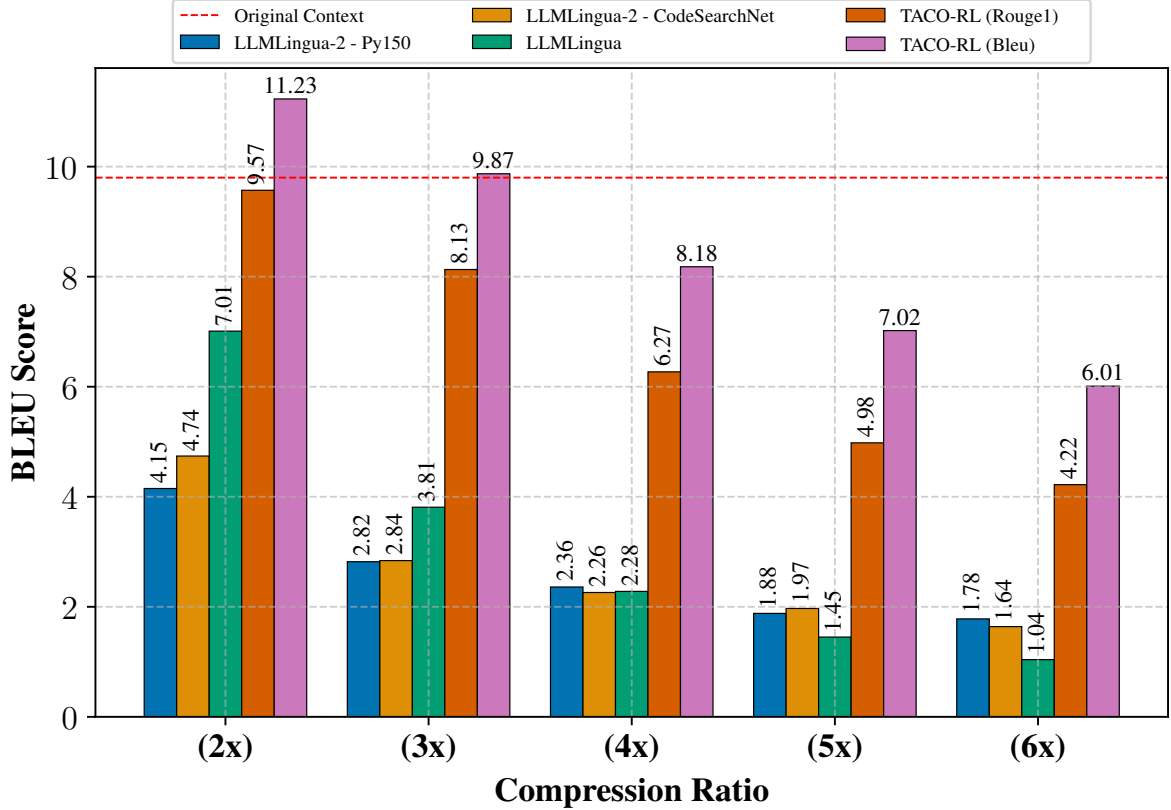}}
  \caption{Comparison of BLEU score across different compression rates for various models on the \textbf{CodeSearchNet} dataset.}
  \label{fig:csn_scores}
\end{figure*}

\textbf{MeetingBank:} Table \ref{meetingbank_results} presents the results for the meeting text summarization task. We report BLEU, ROUGE, and BERT Score \cite{Zhang*2020BERTScore}. Our approach yields a significant performance improvement, with a performance boost of over \textbf{14\%} in BLEU scores at \textit{2x} compression, which further increases to \textbf{45\%} at \textit{6x} compression. We observe similar trends for other metrics. Notably, at \textit{2x} compression, our model performance is close to the results with using the original contexts.

\textbf{SQuAD 2.0:}  Figure \ref{fig:squad_scores} shows the trend of the model performance with increasing compression ratios on the QA task. More detailed results are provided in Table \ref{squad_scores} in the appendix. Our method outperforms the LLMLingua-2 baseline by \textbf{11\%} and \textbf{22\%} at \textit{2x} compression and \textbf{43\%} and \textbf{63\%} at \textit{6x} compression in F1 score and Exact Match score respectively.

\textbf{CodeSearchNet:} Figure \ref{fig:csn_scores} shows the results for the code summarization task. Even though the base model performs poorly, leveraging our approach, the final fine-tuned model significantly outperforms all other baselines. Our method shows gains ranging from \textbf{0.91x (91\%) at \textit{2x} compression to \textbf{1.89x} at \textit{5x}} compression in BLEU scores compared to the LLMLingua-2 baseline. See Table \ref{codesearchnet_scores} in the appendix for detailed results. Additionally, Appendix \ref{appendix:E} highlights the differences in the compression mechanisms between TACO-RL and LLMLingua-2.

\subsection{Ablation Study}

We performed an ablation study to assess the impact of different reward metrics on model performance. For the text summarization task, we compared the following metrics: ROUGE-1, ROUGE-L, and BLEU, as shown in Table \ref{squad_ablation_scores}. For a detailed comparison over various compression ratios, see Table \ref{meetingbank_ablation_results_complete}. For the QA task, we evaluated three reward functions: F1 score, token-wise similarity, and a combination of both, as shown in Table \ref{squad_ablation_scores} (with additional details in Table \ref{squad_scores}). The token-wise similarity score is based on token overlaps between the question and context sentences, with a detailed explanation provided in Appendix \ref{appendix:C}. Additionally, we investigate the effect of the entropy regularization term $H(\mathbf{p})$ on the fine-tuned model's performance on the MeetingBank dataset, as detailed in Table \ref{tab:with_without_entropy}.

\section{Conclusion}
In this paper, we introduce TACO-RL, a novel reinforcement learning-based prompt compression method to address computational challenges in large language models (LLMs). Our approach leverages task-specific reward signals to fine-tune a Transformer encoder-based compression model using on-policy RL, enabling effective compression while maintaining performance. Our approach performs consistently well across high compression ratios (up to 6x) without requiring re-training.
We demonstrate significant performance improvements of \textbf{45\%}, \textbf{63\%}, and up to \textbf{1.89x} on the \textit{Text Summarization}, \textit{QA}, and \textit{Code Summarization} tasks respectively, compared to strong baselines. 
%By applying task-specific compression, our approach can reduce the computational cost of LLMs while preserving much of the original capabilities.
% Our work has important implications for deploying LLMs in real-world applications. It provides a promising approach for optimizing prompt compression, paving the way for more efficient and performant NLP systems.
Our work offers a promising approach to optimizing prompt compression,  paving the way for more efficient and performant NLP systems.

\section*{Limitations}
Our approach to prompt compression using reinforcement learning and encoder models has few limitations. Firstly, the fine-tuning process is sensitive to the choice of reward function, base model, and task-specific prompts. Experiments with different reward metrics, such as BLEU and Rouge in Tables \ref{squad_ablation_scores}, \ref{codesearchnet_scores}, \ref{meetingbank_ablation_results_complete}, show that the choice of metric can significantly impact the performance of the encoder model. Additionally, the quality of the base model, which is trained on a similar dataset, can significantly affect the fine-tuning results.

Secondly, the intricacies of task complexity and dataset magnitude significantly influence the fine-tuning methodology. As illustrated in Figure \ref{fig:csn_scores}, fine-tuned models demonstrate enhanced performance on code datasets, yet this improvement diminishes markedly when transitioning from code summarization to code-completion tasks (See Appendix \ref{appendix:C}). Extensive documentation strings within code contexts can potentially skew the compression dynamics, as models might develop strategies to retain doc-string tokens as a form of reward manipulation. Moreover, comprehensive and expansive datasets are essential for effectively capturing and learning the nuanced complexities, particularly in reinforcement learning scenarios.

Finally, the fine-tuning process is computationally expensive, requiring significant resources and time. Optimizing the computational efficiency of the fine-tuning process is an important consideration for future work.
Addressing these limitations will be crucial for improving the practicality and scalability of our approach.

\bibliography{references}

\begin{thebibliography}{36}
\providecommand{\natexlab}[1]{#1}

\bibitem[{Achiam et~al.(2023)Achiam, Adler, Agarwal, Ahmad, Akkaya, Aleman, Almeida, Altenschmidt, Altman, Anadkat et~al.}]{achiam2023gpt}
Josh Achiam, Steven Adler, Sandhini Agarwal, Lama Ahmad, Ilge Akkaya, Florencia~Leoni Aleman, Diogo Almeida, Janko Altenschmidt, Sam Altman, Shyamal Anadkat, et~al. 2023.
\newblock Gpt-4 technical report.
\newblock \emph{arXiv preprint arXiv:2303.08774}.

\bibitem[{Ahmed et~al.(2023)Ahmed, Ghosh, Bansal, Zimmermann, Zhang, and Rajmohan}]{ahmed2023recommending}
Toufique Ahmed, Supriyo Ghosh, Chetan Bansal, Thomas Zimmermann, Xuchao Zhang, and Saravan Rajmohan. 2023.
\newblock Recommending root-cause and mitigation steps for cloud incidents using large language models.
\newblock In \emph{2023 IEEE/ACM 45th International Conference on Software Engineering (ICSE)}, pages 1737--1749. IEEE.

\bibitem[{Brown et~al.(2020)Brown, Mann, Ryder, Subbiah, Kaplan, Dhariwal, Neelakantan et~al.}]{brown2020languagemodelsfewshotlearners}
Tom~B. Brown, Benjamin Mann, Nick Ryder, Melanie Subbiah, Jared Kaplan, Prafulla Dhariwal, Arvind Neelakantan, et~al. 2020.
\newblock \href {https://arxiv.org/abs/2005.14165} {Language models are few-shot learners}.
\newblock \emph{Preprint}, arXiv:2005.14165.

\bibitem[{Chen et~al.(2021)Chen, Tworek, Jun, Yuan, Pinto, Kaplan, Edwards, Burda, Joseph, Brockman et~al.}]{chen2021evaluating}
Mark Chen, Jerry Tworek, Heewoo Jun, Qiming Yuan, Henrique Ponde De~Oliveira Pinto, Jared Kaplan, Harri Edwards, Yuri Burda, Nicholas Joseph, Greg Brockman, et~al. 2021.
\newblock Evaluating large language models trained on code.
\newblock \emph{arXiv preprint arXiv:2107.03374}.

\bibitem[{Chowdhery et~al.(2023)Chowdhery, Narang, Devlin, Bosma, Mishra, Roberts, Barham, Chung, Sutton, Gehrmann et~al.}]{chowdhery2022palmscalinglanguagemodeling}
Aakanksha Chowdhery, Sharan Narang, Jacob Devlin, Maarten Bosma, Gaurav Mishra, Adam Roberts, Paul Barham, Hyung~Won Chung, Charles Sutton, Sebastian Gehrmann, et~al. 2023.
\newblock Palm: Scaling language modeling with pathways.
\newblock \emph{Journal of Machine Learning Research}, 24(240):1--113.

\bibitem[{Chuang et~al.(2024)Chuang, Xing, Chang, Liu, Chen, and Hu}]{chuang2024learningcompresspromptnatural}
Yu-Neng Chuang, Tianwei Xing, Chia-Yuan Chang, Zirui Liu, Xun Chen, and Xia Hu. 2024.
\newblock \href {https://arxiv.org/abs/2402.18700} {Learning to compress prompt in natural language formats}.
\newblock \emph{Preprint}, arXiv:2402.18700.

\bibitem[{Conneau et~al.(2020)Conneau, Khandelwal, Goyal, Chaudhary, Wenzek, Guzm{\'a}n, Grave, Ott, Zettlemoyer, and Stoyanov}]{conneau-etal-2020-unsupervised}
Alexis Conneau, Kartikay Khandelwal, Naman Goyal, Vishrav Chaudhary, Guillaume Wenzek, Francisco Guzm{\'a}n, Edouard Grave, Myle Ott, Luke Zettlemoyer, and Veselin Stoyanov. 2020.
\newblock \href {https://doi.org/10.18653/v1/2020.acl-main.747} {Unsupervised cross-lingual representation learning at scale}.
\newblock In \emph{Proceedings of the 58th Annual Meeting of the Association for Computational Linguistics}, pages 8440--8451, Online. Association for Computational Linguistics.

\bibitem[{Devlin et~al.(2019)Devlin, Chang, Lee, and Toutanova}]{devlin2019bertpretrainingdeepbidirectional}
Jacob Devlin, Ming-Wei Chang, Kenton Lee, and Kristina Toutanova. 2019.
\newblock \href {https://arxiv.org/abs/1810.04805} {Bert: Pre-training of deep bidirectional transformers for language understanding}.
\newblock \emph{Preprint}, arXiv:1810.04805.

\bibitem[{Goel et~al.(2024)Goel, Husain, Singh, Ghosh, Parayil, Bansal, Zhang, and Rajmohan}]{goel2024x}
Drishti Goel, Fiza Husain, Aditya Singh, Supriyo Ghosh, Anjaly Parayil, Chetan Bansal, Xuchao Zhang, and Saravan Rajmohan. 2024.
\newblock X-lifecycle learning for cloud incident management using llms.
\newblock In \emph{Companion Proceedings of the 32nd ACM International Conference on the Foundations of Software Engineering}, pages 417--428.

\bibitem[{Hu et~al.(2023)Hu, Ganter, Deilamsalehy, Dernoncourt, Foroosh, and Liu}]{hu2023meetingbankbenchmarkdatasetmeeting}
Yebowen Hu, Tim Ganter, Hanieh Deilamsalehy, Franck Dernoncourt, Hassan Foroosh, and Fei Liu. 2023.
\newblock \href {https://arxiv.org/abs/2305.17529} {Meetingbank: A benchmark dataset for meeting summarization}.
\newblock \emph{Preprint}, arXiv:2305.17529.

\bibitem[{Huang et~al.(2024)Huang, Zhang, Cheng, Yang, and Yang}]{huang2024fewermoreboostingllm}
Xijie Huang, Li~Lyna Zhang, Kwang-Ting Cheng, Fan Yang, and Mao Yang. 2024.
\newblock \href {https://arxiv.org/abs/2312.08901} {Fewer is more: Boosting llm reasoning with reinforced context pruning}.
\newblock \emph{Preprint}, arXiv:2312.08901.

\bibitem[{Husain et~al.(2020)Husain, Wu, Gazit, Allamanis, and Brockschmidt}]{husain2020codesearchnetchallengeevaluatingstate}
Hamel Husain, Ho-Hsiang Wu, Tiferet Gazit, Miltiadis Allamanis, and Marc Brockschmidt. 2020.
\newblock \href {https://arxiv.org/abs/1909.09436} {Codesearchnet challenge: Evaluating the state of semantic code search}.
\newblock \emph{Preprint}, arXiv:1909.09436.

\bibitem[{Jiang et~al.(2023)Jiang, Wu, Lin, Yang, and Qiu}]{jiang-etal-2023-llmlingua}
Huiqiang Jiang, Qianhui Wu, Chin-Yew Lin, Yuqing Yang, and Lili Qiu. 2023.
\newblock \href {https://doi.org/10.18653/v1/2023.emnlp-main.825} {{LLML}ingua: Compressing prompts for accelerated inference of large language models}.
\newblock In \emph{Proceedings of the 2023 Conference on Empirical Methods in Natural Language Processing}, pages 13358--13376, Singapore. Association for Computational Linguistics.

\bibitem[{Jiang et~al.(2024)Jiang, Wu, Luo, Li, Lin, Yang, and Qiu}]{jiang2024longllmlinguaacceleratingenhancingllms}
Huiqiang Jiang, Qianhui Wu, Xufang Luo, Dongsheng Li, Chin-Yew Lin, Yuqing Yang, and Lili Qiu. 2024.
\newblock \href {https://arxiv.org/abs/2310.06839} {Longllmlingua: Accelerating and enhancing llms in long context scenarios via prompt compression}.
\newblock \emph{Preprint}, arXiv:2310.06839.

\bibitem[{Jung and Kim(2024)}]{Jung_2024}
Hoyoun Jung and Kyung-Joong Kim. 2024.
\newblock \href {https://doi.org/10.1109/access.2024.3403426} {Discrete prompt compression with reinforcement learning}.
\newblock \emph{IEEE Access}, 12:72578–72587.

\bibitem[{Lewis et~al.(2021)Lewis, Perez, Piktus, Petroni, Karpukhin, Goyal, Küttler, Lewis, tau Yih, Rocktäschel, Riedel, and Kiela}]{lewis2021retrievalaugmentedgenerationknowledgeintensivenlp}
Patrick Lewis, Ethan Perez, Aleksandra Piktus, Fabio Petroni, Vladimir Karpukhin, Naman Goyal, Heinrich Küttler, Mike Lewis, Wen tau Yih, Tim Rocktäschel, Sebastian Riedel, and Douwe Kiela. 2021.
\newblock \href {https://arxiv.org/abs/2005.11401} {Retrieval-augmented generation for knowledge-intensive nlp tasks}.
\newblock \emph{Preprint}, arXiv:2005.11401.

\bibitem[{Li et~al.(2023)Li, Dong, Guerin, and Lin}]{li-etal-2023-compressing}
Yucheng Li, Bo~Dong, Frank Guerin, and Chenghua Lin. 2023.
\newblock \href {https://doi.org/10.18653/v1/2023.emnlp-main.391} {Compressing context to enhance inference efficiency of large language models}.
\newblock In \emph{Proceedings of the 2023 Conference on Empirical Methods in Natural Language Processing}, pages 6342--6353, Singapore. Association for Computational Linguistics.

\bibitem[{Loshchilov and Hutter(2017)}]{loshchilov2017sgdrstochasticgradientdescent}
Ilya Loshchilov and Frank Hutter. 2017.
\newblock \href {https://arxiv.org/abs/1608.03983} {Sgdr: Stochastic gradient descent with warm restarts}.
\newblock \emph{Preprint}, arXiv:1608.03983.

\bibitem[{Merity et~al.(2016)Merity, Xiong, Bradbury, and Socher}]{merity2016pointersentinelmixturemodels}
Stephen Merity, Caiming Xiong, James Bradbury, and Richard Socher. 2016.
\newblock \href {https://arxiv.org/abs/1609.07843} {Pointer sentinel mixture models}.
\newblock \emph{Preprint}, arXiv:1609.07843.

\bibitem[{Mu et~al.(2024)Mu, Li, and Goodman}]{mu2024learning}
Jesse Mu, Xiang~Lisa Li, and Noah Goodman. 2024.
\newblock Learning to compress prompts with gist tokens.
\newblock In \emph{Proceedings of the 37th International Conference on Neural Information Processing Systems}, NIPS '23, Red Hook, NY, USA. Curran Associates Inc.

\bibitem[{Ouyang et~al.(2022)Ouyang, Wu, Jiang, Almeida, Wainwright, Mishkin, Zhang, Agarwal, Slama, Ray et~al.}]{ouyang2022training}
Long Ouyang, Jeffrey Wu, Xu~Jiang, Diogo Almeida, Carroll Wainwright, Pamela Mishkin, Chong Zhang, Sandhini Agarwal, Katarina Slama, Alex Ray, et~al. 2022.
\newblock Training language models to follow instructions with human feedback.
\newblock \emph{Advances in neural information processing systems}, 35:27730--27744.

\bibitem[{Pan et~al.(2024)Pan, Wu, Jiang, Xia, Luo, Zhang, Lin, Rühle, Yang, Lin, Zhao, Qiu, and Zhang}]{pan2024llmlingua2datadistillationefficient}
Zhuoshi Pan, Qianhui Wu, Huiqiang Jiang, Menglin Xia, Xufang Luo, Jue Zhang, Qingwei Lin, Victor Rühle, Yuqing Yang, Chin-Yew Lin, H.~Vicky Zhao, Lili Qiu, and Dongmei Zhang. 2024.
\newblock \href {https://arxiv.org/abs/2403.12968} {Llmlingua-2: Data distillation for efficient and faithful task-agnostic prompt compression}.
\newblock \emph{Preprint}, arXiv:2403.12968.

\bibitem[{Papineni et~al.(2002)Papineni, Roukos, Ward, and Zhu}]{papineni-etal-2002-bleu}
Kishore Papineni, Salim Roukos, Todd Ward, and Wei-Jing Zhu. 2002.
\newblock \href {https://doi.org/10.3115/1073083.1073135} {{B}leu: a method for automatic evaluation of machine translation}.
\newblock In \emph{Proceedings of the 40th Annual Meeting of the Association for Computational Linguistics}, pages 311--318, Philadelphia, Pennsylvania, USA. Association for Computational Linguistics.

\bibitem[{Rajpurkar et~al.(2016)Rajpurkar, Zhang, Lopyrev, and Liang}]{rajpurkar2016squad100000questionsmachine}
Pranav Rajpurkar, Jian Zhang, Konstantin Lopyrev, and Percy Liang. 2016.
\newblock \href {https://arxiv.org/abs/1606.05250} {Squad: 100,000+ questions for machine comprehension of text}.
\newblock \emph{Preprint}, arXiv:1606.05250.

\bibitem[{Raychev et~al.(2016{\natexlab{a}})Raychev, Bielik, and Vechev}]{10.1145/3022671.2984041}
Veselin Raychev, Pavol Bielik, and Martin Vechev. 2016{\natexlab{a}}.
\newblock \href {https://doi.org/10.1145/3022671.2984041} {Probabilistic model for code with decision trees}.
\newblock \emph{SIGPLAN Not.}, 51(10):731–747.

\bibitem[{Raychev et~al.(2016{\natexlab{b}})Raychev, Bielik, and Vechev}]{10.1145/2983990.2984041}
Veselin Raychev, Pavol Bielik, and Martin Vechev. 2016{\natexlab{b}}.
\newblock \href {https://doi.org/10.1145/2983990.2984041} {Probabilistic model for code with decision trees}.
\newblock In \emph{Proceedings of the 2016 ACM SIGPLAN International Conference on Object-Oriented Programming, Systems, Languages, and Applications}, OOPSLA 2016, page 731–747, New York, NY, USA. Association for Computing Machinery.

\bibitem[{Reimers and Gurevych(2019)}]{reimers2019sentencebertsentenceembeddingsusing}
Nils Reimers and Iryna Gurevych. 2019.
\newblock \href {https://arxiv.org/abs/1908.10084} {Sentence-bert: Sentence embeddings using siamese bert-networks}.
\newblock \emph{Preprint}, arXiv:1908.10084.

\bibitem[{Schulman et~al.(2017)Schulman, Wolski, Dhariwal, Radford, and Klimov}]{schulman2017proximal}
John Schulman, Filip Wolski, Prafulla Dhariwal, Alec Radford, and Oleg Klimov. 2017.
\newblock Proximal policy optimization algorithms.
\newblock \emph{arXiv preprint arXiv:1707.06347}.

\bibitem[{Shannon(1948)}]{6773024}
C.~E. Shannon. 1948.
\newblock \href {https://doi.org/10.1002/j.1538-7305.1948.tb01338.x} {A mathematical theory of communication}.
\newblock \emph{The Bell System Technical Journal}, 27(3):379--423.

\bibitem[{Skalse et~al.(2022)Skalse, Howe, Krasheninnikov, and Krueger}]{skalse2022definingcharacterizingrewardhacking}
Joar Skalse, Nikolaus H.~R. Howe, Dmitrii Krasheninnikov, and David Krueger. 2022.
\newblock \href {https://arxiv.org/abs/2209.13085} {Defining and characterizing reward hacking}.
\newblock \emph{Preprint}, arXiv:2209.13085.

\bibitem[{Ushio et~al.(2023)Ushio, Alva-Manchego, and Camacho-Collados}]{ushio2023empirical}
Asahi Ushio, Fernando Alva-Manchego, and Jose Camacho-Collados. 2023.
\newblock An empirical comparison of lm-based question and answer generation methods.
\newblock \emph{arXiv preprint arXiv:2305.17002}.

\bibitem[{Wei et~al.(2023)Wei, Wang, Schuurmans, Bosma, Ichter, Xia, Chi, Le, and Zhou}]{wei2023chainofthoughtpromptingelicitsreasoning}
Jason Wei, Xuezhi Wang, Dale Schuurmans, Maarten Bosma, Brian Ichter, Fei Xia, Ed~Chi, Quoc Le, and Denny Zhou. 2023.
\newblock \href {https://arxiv.org/abs/2201.11903} {Chain-of-thought prompting elicits reasoning in large language models}.
\newblock \emph{Preprint}, arXiv:2201.11903.

\bibitem[{Williams(1992)}]{williams1992simple}
Ronald~J. Williams. 1992.
\newblock \href {https://doi.org/10.1007/BF00992696} {Simple statistical gradient-following algorithms for connectionist reinforcement learning}.
\newblock \emph{Mach. Learn.}, 8(3–4):229–256.

\bibitem[{Xu et~al.(2024)Xu, Shi, and Choi}]{xu2024recomp}
Fangyuan Xu, Weijia Shi, and Eunsol Choi. 2024.
\newblock \href {https://openreview.net/forum?id=mlJLVigNHp} {{RECOMP}: Improving retrieval-augmented {LM}s with context compression and selective augmentation}.
\newblock In \emph{The Twelfth International Conference on Learning Representations}.

\bibitem[{Zhang* et~al.(2020)Zhang*, Kishore*, Wu*, Weinberger, and Artzi}]{Zhang*2020BERTScore}
Tianyi Zhang*, Varsha Kishore*, Felix Wu*, Kilian~Q. Weinberger, and Yoav Artzi. 2020.
\newblock \href {https://openreview.net/forum?id=SkeHuCVFDr} {Bertscore: Evaluating text generation with bert}.
\newblock In \emph{International Conference on Learning Representations}.

\bibitem[{Zhang et~al.(2024)Zhang, Ghosh, Bansal, Wang, Ma, Kang, and Rajmohan}]{zhang2024automated}
Xuchao Zhang, Supriyo Ghosh, Chetan Bansal, Rujia Wang, Minghua Ma, Yu~Kang, and Saravan Rajmohan. 2024.
\newblock Automated root causing of cloud incidents using in-context learning with gpt-4.
\newblock In \emph{Companion Proceedings of the 32nd ACM International Conference on the Foundations of Software Engineering}, pages 266--277.

\end{thebibliography}

\appendix
\addcontentsline{toc}{section}{Appendices}
% \section*{Appendices}

\section{Details of Experiments}
\label{appendix:A}
All the experiments were conducted on a 8 V100 cluster instance with 32 GB memory each. 

Throughout all the experiments the input prompts were kept at 512 token length due to the fixed input sequence length of the encoder used. For training samples larger that this length, we chunked them into section of 512 tokens and used each chunk separately. For testing we compressed each chunk individually and then concatenated all compressed chunks into a final compressed prompts.

The time duration for the experiments is heavily dependent on the rate limits of the APIs employed, since at each step of training we need to call the API to get outputs from GPT3.5. Although for the Original Prompts the output can be obtained once and then reused for further epochs. We used API endpoints with max 300k TPM (Token Per Minute) limit. Despite the large TPM limit, the inference speed is greatly affected by traffic.

\textbf{MeetingBank} For the base model we used the WikiText dataset trained on roughly 23k examples for 10 epochs. This base model was then fine-tuned on $\sim$44k samples created from the MeetingBank \cite{hu2023meetingbankbenchmarkdatasetmeeting} after chunking, for 4 epochs using our approach. The test set comprised of 862 meeting transcripts. 

For this dataset, we used target (reference) summaries generated using GPT-4 instead of the using the original summaries. This was done to keep the experiments compatible with the LLMLingua \cite{jiang-etal-2023-llmlingua, pan2024llmlingua2datadistillationefficient} baselines.

\textbf{SQuAD 2.0} The base model for this this dataset was the same as used in the MeetingBank dataset due to the similarity in distribution. In both datasets the context consists of general English language, making WikiText a good candidate for the base model.

For the fine-tuning, a subset of the original SQuAD 2.0 \cite{rajpurkar2016squad100000questionsmachine} dataset with $\sim$34k samples was used with 15 epochs of training. This subset was created by removing very long and very short contexts. We also focused on having same contexts with different questions to test out approach more rigorously. The test set contained $\sim$6k examples. During evaluation, original answers present in the dataset were treated as reference answers.

\textbf{CodeSearchNet} We only used the Python subset of the entire corpus in our experiments.
The base model for this dataset was trained on a subset of Py150 \cite{10.1145/2983990.2984041} dataset with $\sim$20k samples for 10 epochs.

We fine-tuned our model on a curated subset of $\sim$25k samples from the CodeSearchNet \cite{husain2020codesearchnetchallengeevaluatingstate} dataset over 4 epochs. Recognizing the unique challenges in code context summarization, we preprocessed the dataset by eliminating extremely long and short code contexts. This strategic filtering enables the model to focus on learning the most informative sections of code contexts, rather than getting distracted by peripheral tokens or overly verbose implementations.

The evaluation was conducted on a test subset comprising $\sim$1300 Python code examples. Due to the lack of usable summaries in the dataset, reference summaries were generated using complete original prompts with GPT 3.5.

Table \ref{tab:hyper-parameters} shows the various hyper-parameter values used across the experiments:

\begin{flushleft}
\begin{minipage}{\columnwidth}
\begin{table}[H]
\centering
\renewcommand{\arraystretch}{2}
\resizebox{\columnwidth}{!}{
\begin{tabular}{|c|c|c|c|c|c|c|}
\hline
\textbf{Dataset} & \textbf{Epochs} & \textbf{LR} & \textbf{c} & \textbf{L} & \textbf{$\lambda$} & \textbf{$r_{0}$} \\
\hline
MeetingBank & 4 & $1e^{-6}$ & 0.5 & 30 & 0.01 & -0.1 \\
\hline
SQuAD 2.0 & 15 & $1e^{-6}$ & 0.5 & 30 & 0.01 & -0.1 \\
\hline
CodeSearchNet & 4 & $1e^{-6}$ & 0.5 & 30 & 0.001 & -0.1 \\
\hline
\end{tabular}
}
\caption{Hyper-parameter choices for the experiments}
\label{tab:hyper-parameters}
\end{table}
\end{minipage}
\end{flushleft}

\vspace{1em}

% \setkeys{Gin}{draft}
% \begin{figure*}[ht]
%   \includegraphics[width=0.48\linewidth]{Figures/compression_flexibility_plot.png}
%   \hfill
%   \includegraphics[width=0.48\linewidth]{Figures/tolerance_threshold_plot.png}
%   \caption{Impact on downstream task performance by different \textit{c}, \textit{L} values during training on the \textbf{MeetingBank} dataset.}
%   \label{fig:c_and_l}
% \end{figure*}

\begin{figure*}[ht]
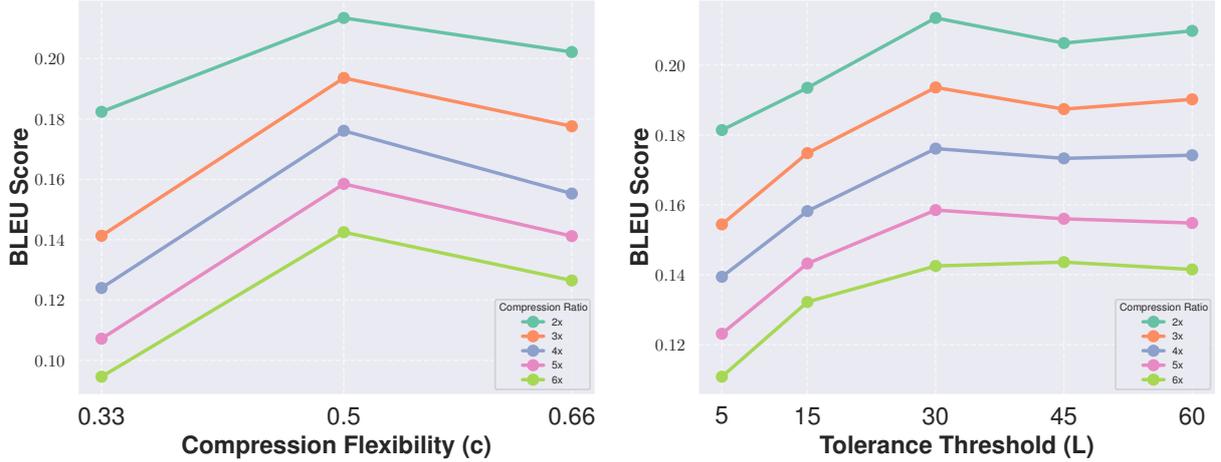

    \scalebox{0.65}{\input{Figures/compression_flexibility_plot.pgf}}
    \scalebox{0.65}{\input{Figures/tolerance_threshold_plot.pgf}}
  \caption{Impact on downstream task performance by different \textit{c}, \textit{L} values during training on the \textbf{MeetingBank} dataset.}
  \label{fig:c_and_l}
\end{figure*}

\section{Impact of hyper-parameters on Training}
\label{appendix:B}
The choice of hyper-parameters is heavily influenced by the datasets. Ours, being a RL based approach, requires tuning of these hyper-parameters for best results. We show the variation in results on the MeetingBank dataset as we tweak there parameters in Figure \ref{fig:c_and_l}.

We notice that for certain values of the hyper-parameters the performance is maximum and then decreases on either side of the value. This effect is more dominant with values of the  \textit{compression flexibility controller \textbf{c}}. With \textit{L}, there is a general trend of increase in performance with increasing hyper-parameter values. 

\section{Code Completion Task}
\label{appendix:C}
We also tried to use the idea of prompt compression for Python Code Completion task. Given a code context, we aimed to discard unnecessary tokens and used the compressed context to predict the last line of the context. Although we achieved improvement over the base model scores, the increase in performance was very subtle and also not of any use. 

For the reward in this task, apart from the F1 score, we used a distance based reward that favoured the tokens close to the end of the context. This was done since the next line of the context depends greatly on the last few lines. We evaluated the performance on the same test set of CodeSearchNet dataset as previously used but with last line of the context removed. This is shown in Table \ref{tab:code_comp_scores}.

\begin{table}[h]
\centering
\renewcommand{\arraystretch}{1.5}
\resizebox{\columnwidth}{!}{%
\begin{tabular}{@{}c|c|c|c|c@{}}
\hline
\textbf{\large Model} & \textbf{\large QA F1} & \textbf{\large Best Sub. EM} & \textbf{\large EM Count} & \textbf{\large EM Score} \\
\hline
{\large Base Model - Py150} & {\large 15.24} & {\large 99.92} & {\large 91.00} & {\large 7.00} \\
\hline
{\large Ours} & {\large 17.73} & {\large 100.00} & {\large 116.00} & {\large 8.92} \\
\hline
\end{tabular}%
}
\caption{Performance of our approach on Code Completion Task}
\label{tab:code_comp_scores}
\end{table}

\section{Token-wise Relevance Reward}
\label{appendix:D}
For the SQuAD QA task, we explored a novel token-wise reward, distinct from the other single-value metric rewards. This approach assigns importance to individual tokens, with higher values indicating better preservation of task-relevant tokens in the compressed prompt. This is in contrast with the earlier reward formulation where a single metric value reward was used. The reward is formulated as follows:

\paragraph{Step 1: Sentence-level Similarity} 
Split the original context $\mathbf{x}$ into $m$ sentences, $\mathbf{x} = \{s_1, s_2, \ldots, s_m\}$. For each sentence $s_i$, compute its similarity with the question $q$:
\begin{equation}
    \text{sim}(s_i, q) = \cos \left( \text{Emb}(s_i), \text{Emb}(q) \right)
\end{equation}
where $\text{Emb}(\cdot)$ denotes sentence embeddings from Sentence Transformers \cite{reimers2019sentencebertsentenceembeddingsusing}.

\paragraph{Step 2: Token-level Similarity Assignment} 
Assign each token $x_j \in s_i$ its sentence's similarity score:
\begin{equation}
    \text{sim}(x_j) = \text{sim}(s_i, q) \quad \text{for } x_j \in s_i
\end{equation}

\paragraph{Step 3: Binary Action Masking} 
Apply the binary action vector $\mathbf{a} = (a_1, a_2, \ldots, a_n)$ to mask token-wise similarity scores:
\begin{equation}
    \mathbf{r}_{\text{masked}} = \mathbf{a} \odot \mathbf{r}_{\text{sim}}
\end{equation}

\paragraph{Step 4: Mean Token-wise Reward Calculation} 
Compute the mean of the masked vector for the final relevance reward $r_{\text{sim}}$:
\begin{equation}
    r_{\text{sim}} = \frac{1}{\sum a_i} \sum_{j=1}^n a_j \cdot \text{sim}(x_j)
\end{equation}

\paragraph{Step 5: Reward Integration} 
The reward can be used standalone or combined with the F1 score:
\begin{equation}
    r_{\text{QA}} = r_{\text{F1}} + \alpha \cdot r_{\text{sim}}
\end{equation}

This combined reward aimed to balance accurate question answering with retention of contextually relevant tokens. However, contrary to expectations, this formulation did not improve performance and even degraded the base model's results, as evidenced in Table \ref{squad_scores}. Moreover, when combined with the F1 score, it diminished the effectiveness of the F1 score as a standalone reward.

\section{Prompt Compression Comparison For Code Summarization}
\label{appendix:E}

\subsection{Original Code Context}
\begin{lstlisting}[language=Python, caption={Original Code Context}]
def yixia_download(url, output_dir = '.', merge = True, info_only = False, **kwargs):
    """wrapper"""
    hostname = urlparse(url).hostname
    if 'n.miaopai.com' == hostname:
        smid = match1(url, r'n\.miaopai\.com/media/([^.]+)')
        miaopai_download_by_smid(smid, output_dir, merge, info_only)
        return
    elif 'miaopai.com' in hostname:  #Miaopai
        yixia_download_by_scid = yixia_miaopai_download
        site_info = "Yixia Miaopai"
        scid = match1(url, r'miaopai\.com/show/channel/([^.]+)\.htm') or \
               match1(url, r'miaopai\.com/show/([^.]+)\.htm') or \
               match1(url, r'm\.miaopai\.com/show/channel/([^.]+)\.htm') or \
               match1(url, r'm\.miaopai\.com/show/channel/([^.]+)')
    elif 'xiaokaxiu.com' in hostname:  #Xiaokaxiu
        yixia_download_by_scid = yixia_xiaokaxiu_download
        site_info = "Yixia Xiaokaxiu"
        if re.match(r'http://v.xiaokaxiu.com/v/.+\.html', url):  #PC
            scid = match1(url, r'http://v.xiaokaxiu.com/v/(.+)\.html')
        elif re.match(r'http://m.xiaokaxiu.com/m/.+\.html', url):  #Mobile
            scid = match1(url, r'http://m.xiaokaxiu.com/m/(.+)\.html')
    else:
        pass
\end{lstlisting}

\vspace{13em}

\subsection{Compressed Context - LLMLingua-2}
\begin{lstlisting}[language=Python, caption={Compressed Code by LLMLingua-2}]
 'n.miaopai
(url[^'
 'miaopai.com'
 Miaopai
 match1(url.com/show/channel/([^.]+)\.htm'
 match1[^
 match1([^.]+)')
 'xiaokaxiu.com'
_download
 "Yixia Xiaokaxiu"
 re.match(r'http://v.xiaokaxiu.com/v/.+\.html',
 match1(url r'http://v.xiaokaxiu/v/(.+).html')
 re.match(r'http://m.xiaokaxiu.com/m/.+\.html'
 match1(url, r'http://m.xiaokaxiu.com/m(
 else pass
\end{lstlisting}

\subsection{Compressed Context - TACO-RL}
\begin{lstlisting}[language=Python, caption={Compressed Code by TACO-RL}]
def yixia_download(url, output_dir = '.', merge = True, info_only = False, **kwargs):
 """wrapper"""
 hostname =
 if 'n.miaopai.com' == hostname:
 smid =
 return
 elif 'miaopai.com' in hostname: #Miaopai
 "Yixia Miaopai"
 scid
 elif 'xiaokaxiu.com'
 Xiaokaxiu
 if re #PC
 scid
 elif
+
 pass
\end{lstlisting}

\begin{table*}
\centering
\resizebox{\textwidth}{!}{
\begin{tabular}{l>{\raggedleft\arraybackslash}c>{\raggedleft\arraybackslash}c>{\raggedleft\arraybackslash}c>{\raggedleft\arraybackslash}c>{\raggedleft\arraybackslash}c}
\toprule
\textbf{Models} & \textbf{QA F1 Score} & \textbf{Best Subspan EM} & \textbf{EM Count} & \textbf{EM Score} \\
\midrule
\multicolumn{5}{c}{\textbf{0.50 (2x compression)}} \\
\hline
LLMLingua-2 - Squad & 62.70 & 99.83  & 2246 & 38.02 \\
LLMLingua-2 - Wikitext & 62.71 (+0.01) & 99.86 (+0.03) & 2231 (-15) & 37.77 (-0.25) \\
LLMLingua & 36.62 (-26.09) & 98.87 (-0.96) & 1108 (-1138) & 18.76 (-19.27) \\
Selective Context & 51.39 (-11.32) & 98.68 (-1.15) & 2042 (-204) & 34.57 (-3.45) \\
TACO-RL - with Token Wise Score & 56.67 (-6.03) & 99.78 (-0.05) & 2076 (-170) & 35.14 (-2.88) \\
TACO-RL - with F1 + Token Wise Score & 68.03 (+5.33) & 99.90 (+0.07) & 2637 (+391) &  44.64 (+6.62) \\
\rowcolor{lightgray} \textbf{TACO-RL - with F1} & \textbf{69.62 (+6.91)} & \textbf{99.92 (+0.08)} & \textbf{2736 (+490)} &  \textbf{46.32 (+8.30)} \\
\hline
\multicolumn{5}{c}{\textbf{0.33 (3x compression)}} \\
\hline
LLMLingua-2 - Squad & 53.11  & 99.58  & 1838 & 31.12 \\
LLMLingua-2 - Wikitext & 52.86 (-0.25) & 99.66 (+0.08) & 1848 (+10) & 31.28 (+0.17) \\
LLMLingua & 30.57 (-22.54) & 97.85 (-1.73) & 974 (-864) & 16.49 (-14.63) \\
Selective Context & 46.37 (-6.75) & 98.26 (-1.32) & 1813 (-25) & 30.69 (-0.42) \\
TACO-RL - with Token Wise Score & 46.25 (-6.86) & 99.31 (-0.27) & 1636 (-202) & 27.70 (-3.42) \\
TACO-RL - with F1 + Token Wise Score & 62.79 (+9.68) & 99.81 (+0.24) & 2412 (+574) & 40.83 (+9.72)\\
\rowcolor{lightgray} \textbf{TACO-RL - with F1} & \textbf{65.84 (+12.73)} & \textbf{99.92 (+0.34)} & \textbf{2592 (+754)} & \textbf{43.88 (+12.76)}\\
\hline
\multicolumn{5}{c}{\textbf{0.25 (4x compression)}} \\
\hline
LLMLingua-2 - Squad & 46.37  & 99.27  & 1548 & 26.21 \\
LLMLingua-2 - Wikitext & 46.15 (-0.21) & 99.61 (+0.34) & 1545 (-3) & 26.16 (-0.05)\\
LLMLingua & 28.27 (-18.10) & 97.58 (-1.69) & 918 (-630) & 15.54 (-10.67) \\
Selective Context & 40.35 (-6.02) & 97.90 (-1.37) & 1555 (+7) & 26.32 (+0.12) \\
TACO-RL - with Token Wise Score & 39.85 (-6.51) & 99.17 (-0.10) & 1342 (-206) & 22.72 (-3.49) \\
TACO-RL - with F1 + Token Wise Score & 58.37 (+12.01) & 99.71 (+0.44) & 2195 (+647) & 37.16 (+10.95)\\
\rowcolor{lightgray} \textbf{TACO-RL - with F1} & \textbf{61.70 (+15.33)} & \textbf{99.88 (+0.61)} & \textbf{2386 (+838)} & \textbf{40.39 (+14.19)}\\
\hline
\multicolumn{5}{c}{\textbf{0.20 (5x compression)}} \\
\hline
LLMLingua-2 - Squad & 41.97  & 99.12  & 1391 & 23.55\\
LLMLingua-2 - Wikitext & 41.49 (-0.48) & 99.41 (+0.29) & 1363 (-28) & 23.07 (-0.47)\\
LLMLingua & 26.96 (-15.01) & 97.26 (-1.86) & 866 (-525) & 14.66 (-8.89) \\
Selective Context & 37.37 (-4.61) & 97.83 (-1.29) & 1417 (+26) & 23.99 (+0.44) \\
TACO-RL - with Token Wise Score & 35.20 (-6.77) & 98.97 (-0.15) & 1167 (-224) & 19.76 (-3.79) \\
TACO-RL - with F1 + Token Wise Score & 54.57 (+12.59) & 99.78 (+0.66) & 2024 (+633) & 34.26 (+10.72)\\
\rowcolor{lightgray} \textbf{TACO-RL - with F1} & \textbf{57.92 (+15.95)} & \textbf{99.76 (+0.64)} & \textbf{2199 (+808)} & \textbf{37.23 (+13.68)}\\
\hline
\multicolumn{5}{c}{\textbf{0.166 (6x compression)}} \\
\hline
LLMLingua-2 - Squad & 38.74 & 98.98 & 1268 & 21.47\\
LLMLingua -2 - Wikitext & 38.78 (+0.04) & 99.27 (+0.29) & 1259 (-9) & 21.31 (-0.15)\\
LLMLingua & 25.96 (-12.78) & 97.14 (-1.84) & 848.00 (-420) & 14.36 (-7.11) \\
Selective Context & 35.53 (-3.21) & 97.82 (-1.17) & 1323.00 (+55) & 22.40 (+0.93) \\
TACO-RL - with Token Wise Score & 31.98 (-6.76) & 98.93 (-0.05) & 1011 (-257) & 17.12 (-4.35) \\
TACO-RL - with F1 + Token Wise Score & 51.56 (+12.82) & 99.64 (+0.66) & 1867 (+599) & 31.61 (10.14)\\
\rowcolor{lightgray} \textbf{TACO-RL - with F1} & \textbf{55.46 (+16.72)} & \textbf{99.81 (+0.83)} & \textbf{2067 (+799)} & \textbf{34.99 (13.53)}\\
\hline
\textbf{Results with Original Prompts} & \textbf{71.40} & \textbf{99.93} & \textbf{2805} & \textbf{47.49} \\
\bottomrule
\end{tabular}
}
\caption{\label{squad_scores} Performance metrics for different models across various compression rates on the \textbf{Squad Dataset}. Values in parentheses indicate deltas from the LLMLingua-2 baseline model trained on Squad dataset.}
\end{table*}

\begin{table*}
\centering
\resizebox{\textwidth}{!}{
\begin{tabular}{l>{\raggedleft\arraybackslash}c>{\raggedleft\arraybackslash}c>{\raggedleft\arraybackslash}c>{\raggedleft\arraybackslash}c>{\raggedleft\arraybackslash}c>{\raggedleft\arraybackslash}c}
\toprule
\textbf{Models}  & \textbf{Bleu} & \textbf{Rouge1} & \textbf{Rouge2} & \textbf{RougeL} & \textbf{BertScore F1} \\
\midrule
\multicolumn{6}{c}{\textbf{0.50 (2x compression)}} \\
\hline
LLMLingua-2 - CodeSearchNet  & 18.30 & 51.57 & 22.50 & 37.94 & 90.27 \\
LLMLingua-2 - Py150  & 19.90 (+1.60) & 52.93 (+1.36) & 24.35 (+1.85) & 39.65 (+1.71) & 90.61 (+0.34) \\
LLMLingua & 23.58(+5.28) & 55.23 (+3.66) & 28.64 (+6.14) & 43.41 (+5.47) & 91.12 (+0.85) \\
Selective Context & - & - & - & - & - \\
\textbf{TACO-RL - with Rouge1} & 29.97 (+11.67) & 59.32 (+7.75) & 33.99 (+11.49) & 47.45 (+9.51) & 91.76 (+1.49) \\
\rowcolor{lightgray} \textbf{TACO-RL - with Bleu} & 35.04 (+16.74) & 61.26 (+9.69) & 38.83 (+16.34) & 50.77 (+12.84) & 92.26 (+1.98) \\
\hline

\multicolumn{6}{c}{\textbf{0.33 (3x compression)}} \\
\hline
LLMLingua-2 - CodeSearchNet  & 11.98 & 44.95 & 15.83 & 32.07 & 89.06 \\
LLMLingua-2 - Py150  & 12.11 (+0.14) & 44.94 (-0.01) & 15.81 (-0.02) & 32.10 (+0.02) & 89.10 (+0.04) \\
LLMLingua & 13.40 (+1.42) & 46.29 (+1.34) & 18.21 (+2.40) & 34.49 (+2.42) & 89.33 (+0.28) \\
Selective Context & - & - & - & - & - \\
\textbf{TACO-RL - with Rouge1} & 22.77 (+10.79) & 54.39 (+9.44) & 27.34 (+11.50) & 41.81 (+9.74) & 90.83 (+1.77) \\
\rowcolor{lightgray} \textbf{TACO-RL - with Bleu} & 28.81 (+16.84) & 57.32 (+12.37) & 33.99 (+18.16) & 46.90 (+14.82) & 91.56 (+2.50) \\
\hline

\multicolumn{6}{c}{\textbf{0.25 (4x compression)}} \\
\hline
LLMLingua-2 - CodeSearchNet  & 8.93 & 41.07 & 12.30 & 28.87 & 88.31 \\
LLMLingua-2 - Py150  & 9.35 (+0.42) & 41.62 (+0.55) & 12.81 (+0.51) & 29.21 (+0.34) & 88.45 (+0.14) \\
LLMLingua & 9.06 (+0.13) & 41.51 (+0.44) & 13.69 (+1.39) & 30.15 (+1.28) & 88.34 (+0.02) \\
Selective Context & - & - & - & - & - \\
\textbf{TACO-RL - with Rouge1}  & 17.59 (+8.66) & 50.87 (+9.80) & 23.03 (+10.73) & 38.48 (+9.61) & 90.22 (+1.91) \\
\rowcolor{lightgray} \textbf{TACO-RL - with Bleu} & 24.23 (+15.30) & 54.89 (+13.82) & 31.26 (+18.96) & 44.61 (+15.74) & 91.14 (+2.83) \\
\hline
\multicolumn{6}{c}{\textbf{0.20 (5x compression)}} \\
\hline
LLMLingua-2 - CodeSearchNet  & 7.31 & 38.88 & 10.56 & 27.21 & 87.87 \\
LLMLingua-2 - Py150 & 7.76 (+0.46) & 39.09 (+0.21) & 11.00 (+0.45) & 27.46 (+0.26) & 88.00 (+0.13) \\
LLMLingua & 6.57 (-0.73) & 38.63 (-0.25) & 11.13 (+0.58) & 27.81 (+0.60) & 87.74 (-0.13) \\
Selective Context & - & - & - & - & - \\
\textbf{TACO-RL - with Rouge1}  & 14.15 (+6.84) & 47.94 (+9.06) & 20.20 (+9.65) & 36.10 (+8.89) & 89.76 (+1.89) \\
\rowcolor{lightgray} \textbf{TACO-RL - with Bleu} & 21.13 (+13.82) & 52.67 (+13.79) & 29.14 (+18.58) & 42.84 (+15.63) & 90.81 (+2.94) \\
\hline
\multicolumn{6}{c}{\textbf{0.166 (6x compression)}} \\
\hline
LLMLingua-2 - CodeSearchNet  & 6.54 & 37.28 & 9.58 & 26.20 & 87.64 \\
LLMLingua-2 - Py150  & 6.79 (+0.25) & 37.63 (+0.35) & 9.95 (+0.37) & 26.49 (+0.29) & 87.69 (+0.05) \\
LLMLingua & 5.31 (-1.23) & 36.62 (-0.66) & 9.76 (+0.18) & 26.38 (+0.18) & 87.36 (-0.29) \\
Selective Context & - & - & - & - & - \\
\textbf{TACO-RL - with Rouge1}  & 12.38 (+5.84) & 46.07 (+8.79) & 18.73 (+9.15) & 34.93 (+8.73) & 89.53 (+1.89) \\
\rowcolor{lightgray} \textbf{TACO-RL - with Bleu}  & 18.46 (+11.92) & 50.99 (+13.71) & 27.33 (+17.75) & 41.34 (+15.14) & 90.52 (+2.88) \\
\hline
\textbf{Results with Original Prompts}  & \textbf{87.89} & \textbf{92.87} & \textbf{88.69} & \textbf{91.25} & \textbf{98.61} \\
\bottomrule
\end{tabular}
}
\caption{\label{codesearchnet_scores} Performance metrics for different models across various compression rates on the \textbf{CodeSearchNet Dataset}. The scores for Selective Context are not added as it struggles to compress code data.}
\end{table*}

%############### ABLATION SCORES ####################

\begin{table*}[t]
  \centering
  \resizebox{\textwidth}{!}{
    \begin{tabular}{l>{\raggedleft\arraybackslash}c>{\raggedleft\arraybackslash}c>{\raggedleft\arraybackslash}c>{\raggedleft\arraybackslash}c>{\raggedleft\arraybackslash}c}
        \toprule
        \textbf{Models} & \textbf{Bleu} & \textbf{Rouge1} & \textbf{Rouge2} & \textbf{RougeL} & \textbf{BertScore F1} \\
        \midrule
        \multicolumn{6}{c}{\textbf{0.50 (2x compression)}} \\
        \hline
        LLMLingua-2 - MeetingBank & 18.68 & 54.20 & 29.45 & 40.14 & 90.69 \\
        Ours - with Rouge1 & 19.89 (+1.21) & 54.40 (+0.20) & 30.55 (+1.11) & 41.22 (+1.08) & 90.82 (+0.13) \\
        Ours - with RougeL & 19.72 (+1.04) & 54.11 (-0.10) & 30.56 (+1.12) & 41.20 (+1.06) & 90.79 (+0.10) \\
        \rowcolor{lightgray} \textbf{TACO-RL - with Bleu} & \textbf{21.35 (+2.67)} & \textbf{55.34 (+1.14)} & \textbf{31.88 (+2.43)} & \textbf{42.17 (+2.03)} & \textbf{90.95 (+0.26)} \\
        \hline
        \multicolumn{6}{c}{\textbf{0.33 (3x compression)}} \\
        \hline
        LLMLingua-2 - MeetingBank & 15.11 & 51.67 & 25.60 & 37.18 & 90.17 \\
        TACO-RL - with Rouge1 & 17.40 (+2.29) & 52.50 (+0.83) & 27.39 (+1.79) & 38.75 (+1.57) & 90.34 (+0.18) \\
        TACO-RL - with RougeL & 17.64 (+2.53) & 52.48 (+0.81) & 27.90 (+2.30) & 39.03 (+1.85) & 90.38 (+0.21) \\
        \rowcolor{lightgray} \textbf{TACO-RL - with Bleu} & \textbf{19.36 (+4.26)} & \textbf{53.67 (+1.99)} & \textbf{29.54 (+3.94)} & \textbf{40.01 (+2.83)} & \textbf{90.54 (+0.37)} \\
        \hline
        \multicolumn{6}{c}{\textbf{0.25 (4x compression)}} \\
        \hline
        LLMLingua-2 - MeetingBank & 12.80 & 49.40 & 22.77 & 34.77 & 89.78 \\
        TACO-RL - with Rouge1 & 15.73 (+2.93) & 51.01 (+1.60) & 25.53 (+2.76) & 36.90 (+2.13) & 90.02 (+0.24) \\
        TACO-RL - with RougeL & 15.68 (+2.88) & 50.77 (+1.36) & 25.82 (+3.05) & 37.04 (+2.27) & 90.03 (+0.25) \\
        \rowcolor{lightgray} \textbf{TACO-RL - with Bleu} & \textbf{17.61 (+4.81)} & \textbf{52.33 (+2.92)} & \textbf{27.84 (+5.07)} & \textbf{38.57 (+3.79)} & \textbf{90.26 (+0.48)} \\
        \hline
        \multicolumn{6}{c}{\textbf{0.20 (5x compression)}} \\
        \hline
        LLMLingua-2 - MeetingBank & 11.13 & 47.50 & 21.01 & 33.25 & 89.44 \\
        TACO-RL - with Rouge1 & 13.82 (+2.69) & 49.19 (+1.68) & 23.58 (+2.57) & 35.27 (+2.03) & 89.69 (+0.25) \\
        TACO-RL - with RougeL & 13.98 (+2.86) & 48.73 (+1.23) & 24.10 (+3.09) & 35.35 (+2.10) & 89.71 (+0.28) \\
        \rowcolor{lightgray} \textbf{TACO-RL - with Bleu} & \textbf{15.85 (+4.73)} & \textbf{50.56 (+3.06)} & \textbf{26.04 (+5.03)} & \textbf{36.81 (+3.56)} & \textbf{89.96 (+0.52)} \\
        \hline
        \multicolumn{6}{c}{\textbf{0.166 (6x compression)}} \\
        \hline
        LLMLingua-2 - MeetingBank & 9.80 & 45.82 & 19.19 & 31.64 & 89.12 \\
        TACO-RL - with Rouge1 & 12.78 (+2.98) & 48.08 (+2.26) & 22.49 (+3.30) & 34.24 (+2.61) & 89.49 (+0.36) \\
        TACO-RL - with RougeL & 13.22 (+3.42) & 47.62 (+1.79) & 22.83 (+3.64) & 34.33 (+2.69) & 89.50 (+0.38) \\
        \rowcolor{lightgray} \textbf{TACO-RL - with Bleu} & \textbf{14.25 (+4.45)} & \textbf{48.60 (+2.78)} & \textbf{24.51 (+5.33)} & \textbf{35.08 (+3.44)} & \textbf{89.68 (+0.56)} \\
        \hline
        \textbf{Results with Original Prompts} & \textbf{21.50} & \textbf{55.19} & \textbf{33.03} & \textbf{42.90} & \textbf{91.12} \\
        \bottomrule
        \end{tabular}
    }
    \caption{\label{meetingbank_ablation_results_complete} Performance metrics for different models trained using different rewards across various compression rates on the \textbf{MeetingBank Dataset}. Values in parentheses indicate deltas from the original LLMLingua-2 baseline.
    }
\end{table*}

% ##################################################

% NEW Tables ################################

\begin{table*}[!]
\centering
\begin{tabular}{lcccr}
\toprule
Metric & Mean & Std & 95\% CI & P-value \\
\midrule
BLEU & 21.2461 & 0.0991 & [21.1230, 21.3692] & 1.0000 \\
ROUGE1 & 55.2168 & 0.0728 & [55.1265, 55.3071] & 1.0000 \\
ROUGE2 & 31.7949 & 0.0387 & [31.7469, 31.8430] & 1.0000 \\
ROUGEL & 41.9159 & 0.0309 & [41.8775, 41.9542] & 1.0000 \\
BERTScore & 90.9349 & 0.0060 & [90.9274, 90.9424] & 1.0000 \\
\bottomrule
\end{tabular}
\caption{One-sample t-test results for 2x compression over 5 runs on the \textbf{MeetingBank} Dataset for TACO-RL. The results show mean scores, standard deviation, 95\% confidence intervals, and p-values for various evaluation metrics.}
\label{tab:sig_test_table}
% \end{table*}

\vspace{3em}

% \begin{table*}
% \centering
\begin{tabular}{lccccc}
\toprule
Method & BLEU & ROUGE1 & ROUGE2 & ROUGEL & BERTScore \\
\midrule
LLMlingua-2 - Wikitext & 16.71 & 52.58 & 27.73 & 39.05 & 90.47 \\
Original Prompts & \textbf{21.50} & 55.19 & \textbf{33.03} & \textbf{42.90} & \textbf{91.12} \\
TACO-RL - with entropy term & 21.35 & \textbf{55.34} & 31.88 & 42.17 & 90.95 \\
TACO-RL - without entropy term & 17.62 & 53.39 & 28.59 & 39.49 & 90.57 \\
\bottomrule
\end{tabular}
\caption{Performance comparison of TACO-RL with and without the entropy term $\lambda H(\mathbf{p})$ on \textbf{MeetingBank} dataset at 2x compression ratio. Best results are shown in \textbf{bold}.}
\label{tab:with_without_entropy}
\end{table*}

\end{document}